\documentclass[lettersize,journal]{IEEEtran}
\usepackage[caption=false,font=normalsize,labelfont=sf,textfont=sf]{subfig}
\usepackage{textcomp,verbatim,stfloats,graphicx,cite,array}
\usepackage{amsmath,amssymb,amsfonts}

\usepackage{amsthm}

\usepackage{multicol}

\usepackage{mathtools}
\usepackage[colorlinks=true,allcolors=blue]{hyperref}
\usepackage[dvipsnames]{xcolor}

\usepackage{comment}
\usepackage{algorithm}
\usepackage[noend]{algpseudocode}
 \algrenewcommand\algorithmicindent{0.8em}
\newcommand\algcolor{\color{white!70!black}}
\newcommand\newalgcomment[2]{{\algcolor\hfill~#1 \ref{#2}}}

\usepackage{enumitem}

\makeatletter
\newcommand{\customlabel}[2]{% 
\protected@write\@auxout{}{\string \newlabel{#1}{{#2}{\thepage}{#2}{#1}{}}}% 
\hypertarget{#1}{#2}
}
\makeatother 
\usepackage[many]{tcolorbox}
\tcbuselibrary{breakable}
\makeatletter
\def\tcb@parbox@use@false{% 
  \def\@parboxrestore{\linewidth\hsize\let\@parboxrestore=\tcb@parboxrestore}% 
}

\usepackage{booktabs}
\usepackage{cancel}

\newcommand{\R}{\mathbb{R}}
\newcommand{\diag}{\mathrm{diag}}
\newcommand{\norminf}[1]{\left\lVert #1 \right\rVert_\infty}

\newcommand{\trans}{^\top}
\DeclareMathOperator*{\argmin}{arg\,min}

\newcommand\AverageSmallMatrix[1]{{% 
\footnotesize\arraycolsep=0.22\arraycolsep\ensuremath{\begin{bmatrix}#1\end{bmatrix}}}}

\newcommand\scalemath[2]{\scalebox{#1}{\mbox{\ensuremath{\displaystyle #2}}}}

\usepackage{csquotes}
\usepackage{quoting}

\usepackage{etoolbox}
  
\usepackage{xspace}
\newcommand\OCP{\textbf{OCP}\xspace}

\newcommand\QP{\textbf{QP}\xspace}
\newcommand\ADMM{\texttt{ADMM}\xspace}

\newcommand\OSQP{\texttt{OSQP}\xspace}

\PassOptionsToPackage{dvipsnames}{xcolor}
\usepackage{pifont}
\newcommand{\cmark}{\textcolor{ForestGreen}{\ding{51}}}
\newcommand{\xmark}{\textcolor{BrickRed}{\ding{55}}}

\newcommand{\solverName}[0]{\texttt{TurboMPC}}

\usepackage{cuted}
\usepackage{capt-of}
\title{\solverName: Fast, Scalable, and Differentiable\\Model Predictive Control on the GPU}
\author{Gabriel Bravo-Palacios, % 
        Jianghan Zhang,
        Zachary Pestrikov, % 
        Brian Plancher, and % 
        Thomas Lew% 
\thanks{Gabriel Bravo-Palacios, Jianghan Zhang, Zachary Pestrikov, and Brian Plancher are with Dartmouth College, Hanover, NH, USA (Email: {\tt\scriptsize \{gbravo, jianghan.zhang.gr, zpestrikov, plancher\}@dartmouth.edu})}% 
\thanks{Thomas Lew is with Toyota Research Institute, Los Altos, CA, USA. (Email: {\tt\scriptsize thomas.lew@tri.global})}% 
\thanks{Toyota Research Institute provided funds to support this work, but this article
solely reflects the opinions and conclusions of its authors.
}% 
}

\begin{document}

\maketitle

\begin{strip}
    \centering
    \includegraphics[width=0.95\linewidth]{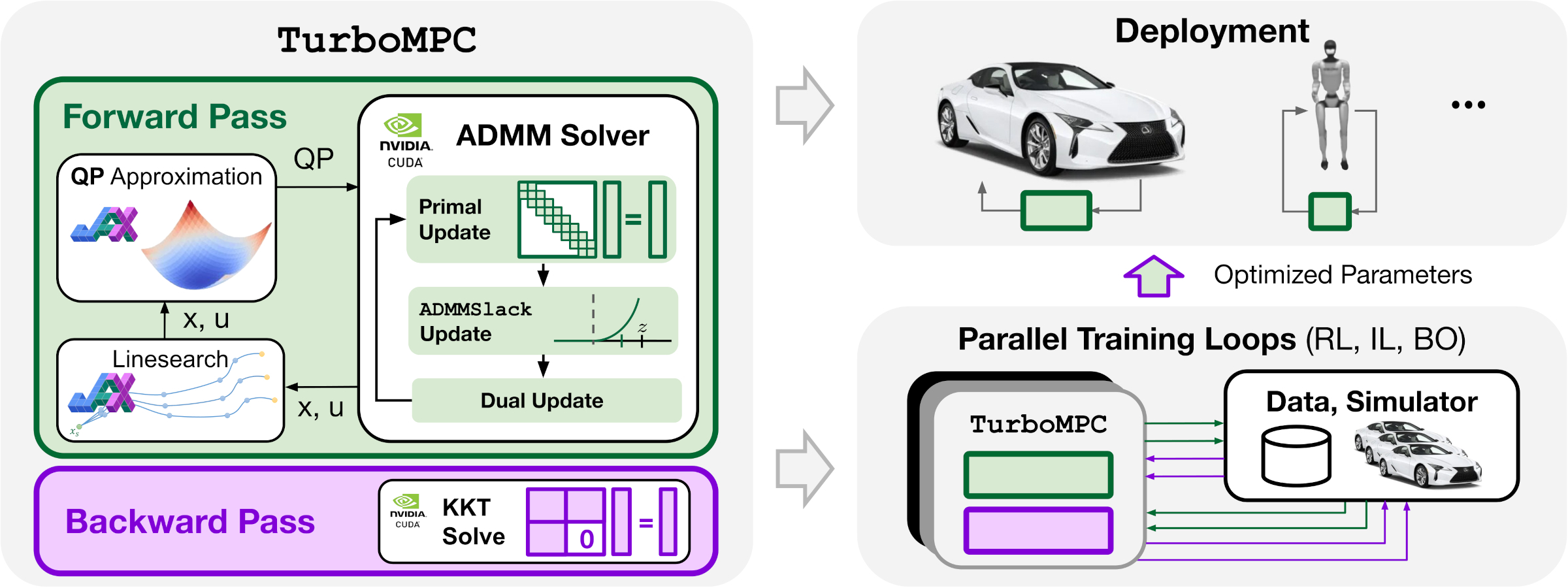}
    \vspace{-5pt}
    \captionof{figure}{\small \solverName~is a differentiable MPC solver that runs entirely on the GPU. It supports realistic problem formulations that enable deployment on challenging robotics applications such as minimum-time racing. Thanks to its differentiability and GPU support, it can be used in learning pipelines such as auto-tuning via reinforcement learning (RL), imitation learning (IL), and Bayesian optimization (BO).}
    \vspace{72pt}
    \label{fig:hero}
\end{strip}

\begin{abstract}
Robotics increasingly relies on GPUs for parallel simulation, large-scale learning, and neural-network inference. For model predictive control (MPC) to scale with this paradigm, 
solvers must run efficiently on this hardware while remaining fast, differentiable, and compatible with expressive MPC formulations used in robotics. 
We present \solverName, a differentiable MPC solver that runs entirely on the GPU and  supports state and control inequality constraints, implicit integrators, cross-time-coupled costs, and slack variables. \solverName{} combines sequential quadratic programming (SQP), an alternating direction method of multipliers (ADMM) inner solver, % 
implicit differentiation, % 
and a co-designed JAX-CUDA implementation for efficiency and ease of use. 
In simulation, we validate \solverName{}  on constrained planning,  humanoid imitation learning, and reinforcement learning with neural-network cost function tasks, % 
achieving up to $15\times$ and $58\times$ speedups over state-of-the-art CPU and GPU  differentiable solvers, respectively. 
We deploy \solverName{} on a full-scale car for minimum-time racing and find that batched, GPU-accelerated tuning of MPC parameters via Bayesian optimization  yields significantly faster driving than a hand-tuned baseline. \solverName{} also scales to planning horizons of over $8000$ knot points while maintaining control of the vehicle. % 
We open-source \solverName{}  at: 

\begin{center}
\url{https://github.com/ToyotaResearchInstitute/turbompc}
\end{center}
\end{abstract}

\newpage
\section{Introduction}\label{sec:intro}
\IEEEPARstart{R}{obotics} 
is increasingly powered by scale: large datasets, parallel simulation, and neural networks trained on  graphics processing units (GPUs)~\cite{zitkovich2023rt,makoviychuk2021isaac,o2024open,rudin2022learning,Barreiros2026}. 
How can optimization-based model predictive control (MPC) scale with this new computational paradigm?

MPC remains one of the most reliable tools for controlling robotic systems, % 
as it enables planning with a model, reasoning about constraints, and reacting to changing conditions~\cite{kuindersma2016optimization,di2018dynamic,mastalli2020motion,tranzatto2022cerberus,wensing2023optimization,grandia2023perceptive,nguyen2024tinympc,bui2025push}. Yet, most high-performance MPC solvers are designed for central processing units (CPUs), where sequential algorithms such as Riccati recursions can exploit the temporal structure of optimal control problems~\cite{Jordana2025,FRISON20206563,Stellato2020,Grandia2023}.
This CPU-centric design limits the integration of MPC into learning-based pipelines that rely on batching, parallel rollouts, and differentiable architectures at scale. % 

While GPU-based differentiable solvers offer a path to scale MPC and bring its structured decision-making into modern learning-based robotics, \emph{useful GPU acceleration is not only a matter of parallel execution}. The solver must also be fast enough for online control and expressive enough to model the dynamics and constraints that determine real-world performance.
MPC formulations may require implicit integration for stiff dynamics, control-rate penalties for input smoothness and actuators protection, state inequalities for safety constraints, and slack variables to maintain feasibility. As shown in Table \ref{tab:feature_coverage}, existing tools force a tradeoff: GPU-based solvers often restrict the problem class to expose parallelism, while CPU-based solvers support richer formulations and online deployment, with limited scalability. % 

\subsection{Related Work}
\paragraph*{Solvers on the CPU}
MPC solvers have long exploited the time-induced sparsity of optimal control problems (\textbf{OCPs}) via sequential Riccati recursions and factorization approaches, achieving efficient resolution on CPUs \cite{Jordana2025,FRISON20206563,Stellato2020,Grandia2023},  
and outperforming general-purpose solvers \cite{wachter2006implementation,gill2005snopt} that do not exploit the \OCP structure. 
In particular, \texttt{acados}~\cite{verschueren2022acados,frey2025differentiable} supports implicit integrators, inequality constraints, % 
and differentiability. 
However, relying on a CPU-based implementation limits scalability to larger models and batch sizes.

\paragraph*{Solvers on the GPU}
GPU-based differentiable solvers for MPC offer a promising path toward scaling and combining optimization-based control and learning. However, existing methods restrict problem classes to leverage both the structure of \textbf{OCPs} and expose parallelism. For example, \texttt{trajax}~\cite{frostig2021trajax}, \texttt{mpc.pytorch}~\cite{amos2018differentiable}, and \texttt{DiffMPC}~\cite{adabag2025differentiable} do not handle implicit integrators, cross-time costs, or state inequality constraints. Other GPU-accelerated solvers, whether gradient-based~\cite{amatucci2025primal,jeon2024cusadi,plancher2018performance,sundaralingam2023curobo,adabag2024mpcgpu,bishop2024relu,chari2024fast,du2026gato,Alboni2026,fang2026safelargescalerobustnonlinear} or sampling-based \cite{williams2017model,alvarez2025real,jardali2025zero}
improve throughput or latency, but either lack differentiability or  do not support the full set of features needed for many   target MPC applications, see Table~\ref{tab:feature_coverage}. 

\paragraph*{Co-Design of Solvers and Learning Infrastructure}
One barrier to GPU-native MPC is the mismatch between learning infrastructure and solver throughput. \texttt{JAX}~\cite{jax2018github} provides automatic differentiation and batching that are used in learning pipelines, but incurs per-kernel launch overhead that is significant for iterative solver design. \texttt{CUDA} provides low-overhead kernels % 
needed for efficient GPU execution, 
but is more difficult to interface with. Moreover, the sparsity structure of the problems in the forward and backward passes of a differentiable MPC solver can be different. These requirements require a careful design to unlock an efficient yet practical GPU-accelerated solver (see Section \ref{sec:solver:design}). % 
Our approach addresses this gap through a \texttt{JAX} frontend and fused \texttt{CUDA} backend, using direct sparse factorization~\cite{cudss} as a shared primitive for online control, batched learning, and differentiation.

\paragraph*{Autonomous Racing}
Driving a vehicle at its limits via MPC remains a challenging problem~\cite{kabzan2020amz,jardali2025zero,LewGreiffICRA2025}, requiring long-horizon planning with track bound constraints, stiff dynamics, control-rate costs to protect actuators, and slack variables to ensure recursive feasibility despite model mismatch. Recent work has integrated learning and automated MPC tuning into racing pipelines through differentiable or gradient-free methods~\cite{frohlich2022contextual,rickenbach2025zipmpc,romero2024actor,Jahncke2026}. However, existing approaches use CPU-only solvers or only partly support GPU acceleration~\cite{LewGreiffICRA2025}, limiting batching and large-scale tuning. 

\begin{table}[t]
\centering
\caption{Solver features: \cmark supported, \xmark not supported, $\boldsymbol{\sim}$ partial.}
\vspace{-8pt}
\label{tab:feature_coverage}
\setlength{\tabcolsep}{3pt}
\renewcommand{\arraystretch}{1.1}
\footnotesize
\resizebox{\columnwidth}{!}{
\begin{tabular}{lcccccc}
\toprule
\textbf{Solver} & \textbf{StateIneq.} & \textbf{Implicit} & \textbf{RateCost} & \textbf{Slack} & \textbf{Diff.} & \textbf{GPU} \\
\midrule
\texttt{acados}~\cite{frey2025differentiable} & \cmark & \cmark & \cmark & \cmark & \cmark & \xmark \\
\begin{tabular}[c]{@{}l@{}}
\texttt{Trajax}~\cite{frostig2021trajax}, \texttt{DiffMPC}~\cite{adabag2025differentiable}\\
\texttt{mpc.pytorch}~\cite{amos2018differentiable}
\end{tabular}
& \xmark & \xmark & \xmark & \xmark & \cmark & \cmark 
\\
\texttt{CusADi}~\cite{jeon2024cusadi}% 
& \cmark & \cmark & \cmark & \cmark  & \xmark & \cmark 
\\
Other GPU Solvers~\cite{amatucci2025primal,williams2016aggressive,Alboni2026,alvarez2025real,plancher2018performance,sundaralingam2023curobo,adabag2024mpcgpu,bishop2024relu,chari2024fast,du2026gato,fang2026safelargescalerobustnonlinear} & $\boldsymbol{\sim}$ & \xmark & \xmark & \xmark & \xmark & \cmark \\
Standard NLP Solvers~\cite{wachter2006implementation,gill2005snopt} & \cmark & \cmark & \cmark & \cmark & \xmark & \xmark \\
\midrule
\textbf{\solverName~(Ours)} & \cmark & \cmark & \cmark & \cmark & \cmark & \cmark \\
\bottomrule
\end{tabular}}
\vspace{-10pt}
\end{table}

\subsection{Contribution}
\label{sec:Contribution}

We propose \solverName, a differentiable GPU-accelerated MPC solver 
that addresses problems of the form 
\begin{tcolorbox}[
title={% 
Optimal Control Problem (\OCP)
},
coltitle=black, 
boxrule=3pt,
colframe=blue!6, % 
enhanced, colback=blue!2, boxsep=0pt, left=5pt, right=5pt]
\vspace{-2mm}
\begin{align*}
\min_{(x,u,\xi)} &
\sum_{t=0}^{N-1}\ell_t
\left(
\AverageSmallMatrix{x_t\\u_t},
\AverageSmallMatrix{x_{t+1}\\u_{t+1}}
\right)
+
\ell_N
\left(
\AverageSmallMatrix{x_N\\u_N}
\right) 
+
\tfrac{\gamma_\xi}{2}
\sum_{t=0}^{N}\|\xi_t\|_2^2\\
\text{s.t.}\ \,  &x_0=x_{\rm init},\\
&f_t(x_t,u_t,x_{t+1},u_{t+1})=0,
&&\hspace{-35mm}t=0,\ldots,N-1,\\
&\underline g_t\le g_t(x_t,u_t)+\delta_\xi\xi_t\le \overline g_t,
&&\hspace{-35mm}t=0,\ldots,N,
\end{align*}
\end{tcolorbox} 
\noindent 
with states $x_t\in\R^{n_x}$, controls $u_t\in\R^{n_u}$, optional equality constraints 
$u_0=u_{\rm init}$ and slack variables $\xi_t$, and horizon $N$. The binary $\delta_\xi\in\{0,1\}$ and the scalar $\gamma_\xi>0$ toggle and penalize $\xi_t$, respectively. The solver supports key features:
\begin{itemize}[leftmargin=5mm]
\item Control rate costs to smooth inputs and protect actuators
\item Implicit integrators to handle stiff dynamics
\item Inequality constraints for obstacle avoidance, actuator limits, and additional constraints
\item Slack variables to improve recursive feasibility
\end{itemize}
Since the solver is differentiable and runs entirely on the GPU, 
it can efficiently:
\begin{itemize}[leftmargin=5mm]
\item plan over long horizons,
\item use expressive costs \& constraints (e.g., neural networks),
\item handle high-dimensional systems, and
\item be used in imitation learning (IL), reinforcement learning (RL), and Bayesian optimization (BO) pipelines.
\end{itemize}
\subsubsection{Approach}
The solver tackles \OCP via sequential quadratic programming (SQP)~\cite{Nocedal2006}. % 
It runs entirely on the GPU and leverages the time-induced sparse structure of \OCP to exploit the GPU's parallel computation capabilities.  The inner quadratic programs (QPs) in the SQP loop are solved by a custom alternating direction method of multipliers (ADMM)  scheme using the Schur complement method. The ADMM scheme is inspired from \OSQP~\cite{Stellato2020}, with the primal-update linear systems solved via \texttt{cuDSS}~\cite{cudss}, and slack variables tackled using ideas from~\cite{LewSlackADMM2025} to simplify the implementation and avoid extending the size of the problem. 
Using implicit differentiation, custom vector-Jacobian products (VJP) give gradients of functions of the \OCP solution by differentiating the active-set KKT conditions and solving a linear system.

\subsubsection{Software}
The solver is implemented in \texttt{JAX} and \texttt{CUDA} to simplify its use and ensure performance. It is open-source: \url{https://github.com/ToyotaResearchInstitute/turbompc}

\subsubsection{Results}
We demonstrate the full feature set above across simulation results and deployment on a Lexus LC500 vehicle. % 
The results show significant speedups over strong baselines such as the CPU solver \texttt{acados}~\cite{frey2025differentiable} and  a hybrid GPU-CPU solver using \texttt{OSQP}~\cite{Stellato2020}. Some highlights of the on-vehicle results include the efficient and automatic tuning of the parameters of \OCP via Bayesian optimization to race  significantly faster, and  real-time re-planning  with horizon $N>8000$, an $8\times$ longer horizon than the largest horizon at which the baseline maintains control of the vehicle.
 
\section{Background}\label{sec:background}

\subsection{Differentiable Model Predictive Control} \label{sec:background:solvers}
MPC \cite{Houska2018} is a planning and control strategy that recursively solves \OCP from the current robot state $x_{\textrm{init}}$ and executes the first control input $u_0$ of the solution $x:=(x_t,u_t)_{t=0}^N$, before replanning again from the next robot state $x_{\textrm{init}}$.  
The \OCP depends on parameters $\theta$ that encode the costs $\ell_t$ (e.g., weights penalizing the distance to the goal) and constraints $f_t,g_t,x_{\textrm{init}}$ (e.g., positions of obstacles or friction coefficients of the vehicle dynamics model). 

A differentiable MPC solver implements two capabilities:

\paragraph*{Forward Pass} $\OCP$ is solved for a given parameter $\theta$ to obtain the solution $x$. The solver design is key to efficiently and reliably solve \OCP, e.g., via an interior-point method or SQP \cite{wachter2006implementation,Nocedal2006,gill2005snopt}.

\paragraph*{Backward Pass}
The sensitivity of the solution with respect to the parameters $\theta$ is computed. One approach is to differentiate through the solver iterations, but this approach can be slow and memory-intensive~ \cite{Shaban2019,Scier_NEURIPS2022}. Instead, we use implicit differentiation at the converged solution \cite{frey2025differentiable,Gould2016,amos2017optnet,amos2018differentiable,Blondel2024,agrawal2020learning}: The primal-dual variables $w=(x,y)$ satisfy the KKT conditions $F(w,\theta)=0$, and sensitivities are obtained by differentiating these conditions. The backward pass then amounts to solving a linear system, which avoids unrolling solver iterates. 

\subsection{Exploiting Time-Induced Sparse Structure}
The finite-horizon structure of \OCP induces sparse linear systems that MPC solvers exploit through Riccati recursions, structured factorizations, or iterative methods~\cite{Jordana2025,FRISON20206563,Stellato2020,song2025parallel,verschueren2022acados,frey2025differentiable,Schwan2023}. For explicit dynamics and per-stage costs, the associated KKT systems are block tridiagonal across time. CPU solvers exploit this structure efficiently for single OCP solves, while GPU methods seek parallelism across time, batches, or sparse linear algebra operations~\cite{frostig2021trajax,amos2018differentiable,adabag2025differentiable,adabag2024mpcgpu,du2026gato,amatucci2025primal,jeon2024cusadi}. PCG-based solvers and parallel cyclic reduction efficiently solve these block-tridiagonal subclass on GPUs~\cite{adabag2024mpcgpu,du2026gato,Jordana2025}, but control-rate costs, implicit dynamics, and general inequalities, essential in robotics applications and autonomous racing~\cite{kabzan2020amz,LewGreiffICRA2025}, introduce cross-time couplings or active-set-dependent KKT structures that can break the aforementioned structure.

Differentiable MPC adds a second structural challenge: the forward and backward passes do not necessarily share the same sparsity structure.  
For the forward pass, using our ADMM approach, the primal linear systems admit a Schur complement reduction to a block-tridiagonal linear system, whereas the linear systems in the backward pass do not admit this reduction. Thus, linear system solver primitives specialized only to the forward block-tridiagonal system \OCP do not directly cover the full differentiable feature set in Table \ref{tab:feature_coverage}. \solverName{} therefore uses \texttt{cuDSS}~\cite{cudss} for GPU-friendly sparse direct factorization as a common linear system solver primitive for the forward and backward passes, supporting implicit dynamics, control-rate costs, inequalities, slack variables, and differentiability in one solver. Other supported (but potentially slower) linear system solvers are listed in Table \ref{tab:backends}. 

\section{Solver Design and Implementation}\label{sec:solver}
The differentiable solver is described in Algorithms \ref{alg:sqp}-\ref{alg:admm}. 
We solve \OCP via SQP~\cite[Chapter~18]{Nocedal2006}.
At each SQP iteration, we form a convex approximation of \OCP  (Section~\ref{sec:solver:qp}). We solve the resulting QP via an ADMM scheme inspired by \texttt{OSQP}~\cite{Stellato2020} and the slack variables handling of~\cite{LewSlackADMM2025} (Section~\ref{subsec:sqp:admm}). Then, the next SQP iterate is selected using a backtracking linesearch (Section~\ref{sec:solver:linesearch}). 
Finally, after convergence of the SQP scheme, gradients can be computed using an active-set VJP  based on implicit differentiation (Section~\ref{subsec:backward}). 
Each component has prior art, but the combined and co-designed implementation is what unlocks the performance of this GPU solver. We discuss design choices and implementation details in Section \ref{sec:solver:design}. For compact presentation, we refer the reader to the Appendix for additional details.

\subsection{\QP Approximation of $\OCP$}\label{sec:solver:qp}

At each SQP iterate, we quadratize  the
cost and linearize the constraints, giving the 
following approximation of \OCP:
\begin{center}
\begin{minipage}{0.73\linewidth}
\begin{tcolorbox}[
title={% 
Quadratic Program (\QP)
},
halign title=center,
coltitle=black, 
boxrule=3pt,
colframe=blue!6, % 
enhanced, colback=blue!2, boxsep=0pt, left=5pt, right=5pt]
\vspace{-2mm}
\begin{align*}
\min_{(x,\xi)}\quad&
\tfrac12 x\trans P x + q\trans x + \tfrac{\gamma_\xi}{2}\|\xi\|^2
\\
\text{s.t.}\quad\,&
C x = c,
\ \ \ 
\underline{G} \le Gx +\delta_\xi\xi
   \le \overline{G},
\end{align*}
\end{tcolorbox}
\end{minipage}
\end{center}
where $x$ 
stacks the  stage variables
$
x:=(x_t,u_t)_{t=0}^N,
$
and $(P,q,C,c, \underline G, G, \overline G)$  are defined in Appendix \ref{apdx:qp}.  

Leveraging the sparsity of \QP enables an efficient solver for three reasons. First, as the cost of \OCP couples only neighboring stages, the matrix $P$ is  block-tridiagonal. Second, the matrix $C$ encodes the linearized dynamics and initial-state equality constraints and is block-bidiagonal. Third, the matrix $G$ encodes the linearized pointwise-in-time inequality constraints and is block diagonal.
\begin{gather}
\begin{split}
&\hspace{2cm}
P=
\AverageSmallMatrix{
D_0 & E_0^\top \\[-2mm]
E_0 & D_1 & \ddots \\[-2mm]
& \ddots & \ddots & E_{N-1}^\top\\
& & E_{N-1} & D_N
},
\\[1mm]
&C=
\AverageSmallMatrix{
A_{\mathrm{init}} & 0 & 0 & \cdots & 0 \\
A_0^{-} & A_0^{+} & 0 & \cdots & 0 \\[-1mm]
\vdots & \ddots & \ddots & \ddots & 0 \\[-1mm]
0 & \cdots & 0 & A_{N-1}^{-} & A_{N-1}^{+}
},
\quad\ \ 
G=
\AverageSmallMatrix{
G_0\ & \  0\      & \ 0 \\
0\   &  \ \ddots\  & \ 0  \\[2mm]
0\   & \   0  \     & \   G_N
}.
\end{split}
\end{gather}

\subsection{Solving \QP via \ADMM}\label{subsec:sqp:admm}
We solve \QP via an ADMM scheme that combines \OSQP's ~\cite{Stellato2020}  splitting scheme and the slack variables handling of~\cite{LewSlackADMM2025}. 
We introduce the slack variables $z$\footnote{The \textit{slack variables} $\xi$ should not be confused with the \textit{slack variables} $z$. The variables $\xi$ relax the inequality constraints. The variables $z$ come from the ADMM scheme and capture the right hand sides of the constraints.} and rewrite \QP as
\begin{align}
\min_{(x,z,\xi)}\ \ &\tfrac12 x\trans P x + q\trans x + \tfrac{\gamma_\xi}{2}\|\xi\|^2\\
\text{s.t.} \ \ \ \, 
&
Ax-z=0,\ \ 
z+\AverageSmallMatrix{0\\\delta_\xi I}\xi
\in[l,u], \nonumber
\end{align}
where $A=\AverageSmallMatrix{
 C \\ G
}$, 
$l=\AverageSmallMatrix{
 c \\ \underline G
}$, and  
$u=\AverageSmallMatrix{
 c \\ \overline G
}$.

Let $y$ be the multipliers associated to the constraints $Ax =
z$. The KKT conditions of \QP define the primal-dual
residuals
\begin{align}\label{eq:admm:residuals}
r_{\mathrm{prim}}=
\norminf{
Ax-z
},\quad r_{\mathrm{dual}}=
\norminf{
\AverageSmallMatrix{
Px+q+A^\top y\\\gamma_\xi\xi+[0;\delta_\xi I]y
}
}.
\end{align}
We terminate when
\begin{subequations}\label{eq:admm:termination:slack}
\begin{align}
r_{\mathrm{prim}}&\le\varepsilon_{\mathrm{abs}}
+\varepsilon_{\mathrm{rel}}\norminf{
(Ax, z)
},
\\
r_{\mathrm{dual}}&\le\varepsilon_{\mathrm{abs}}
+\varepsilon_{\mathrm{rel}}
\norminf{
(Px,A^\top y,q,\gamma_\xi\xi,[0;\delta_\xi I]y)
}.
\end{align}
\end{subequations}

Our ADMM solver is derived as follows. We introduce the duplicated variables $(\tilde x,\tilde z)$ and rewrite  \QP as
\begin{align} \label{eq:admmslack}
\min_{(x,z,\tilde x,\tilde z,\xi)}& \ \ 
\tfrac12 \tilde x^\top P \tilde x + q^\top \tilde x
+ \tfrac{\gamma_\xi}{2}\|\xi\|^2
+ \mathcal{I}_{[l,u]}(z+[0;\delta_\xi I]\xi) \nonumber
\\
\text{s.t.}  \  
\ \, & \ \    
A\tilde x-\tilde z=0,\quad \tilde x-x=0,\quad \tilde z-z=0,
\end{align}
where  $\mathcal{I}_\mathcal{C}$ is the indicator function of a set $\mathcal{C}$, with $\mathcal{I}_{[l,u]}(z):=0$ if $z\in\mathcal{C}$ and $\mathcal{I}_\mathcal{C}(z):=+\infty$ otherwise.

Let \(w\) and \(y\) be the dual variables associated with the constraints \(\tilde x-x=0\) and \(\tilde z-z=0\), respectively. 
The augmented Lagrangian  is then
\begin{align}  \label{eq:admm:aug_Lagrangian}
\mathcal L_{\sigma,\rho}(\tilde x,\tilde z,x,z,\xi,w,y)
&=
\tfrac12 \tilde x^\top P \tilde x + q^\top \tilde x
+ \tfrac{\gamma_\xi}{2}\|\xi\|^2+
\nonumber
\\
&\hspace{-24mm}
+  \mathcal{I}_{[l,u]}(z+[0;\delta_\xi I]\xi)
+ \mathcal{I}_{\{0\}}(A\tilde x-\tilde z)
\\
&\hspace{-24mm}
+ \tfrac{\sigma}{2}\|\tilde x-x+\sigma^{-1}w\|^2
+ \tfrac\rho2\|\tilde z-z+\rho^{-1}y\|^2, \nonumber
\end{align}
where \(\sigma>0\) and \(\rho\succ 0\)  are step-size parameters (with \(\rho\) diagonal).

\textbf{ADMM scheme}: Our QP solver consists of 
1) minimizing $\mathcal L_{\sigma,\rho}$ over $(\tilde x, \tilde z)$, 
2) minimizing $\mathcal L_{\sigma,\rho}$ over  $(x,z,\xi)$, and 
3) updating the multipliers $(w,y)$.  
By also including an over-relaxation strategy, we obtain the following three updates.

\begin{tcolorbox}[
title={% 
\textbf{Remark \customlabel{remark:ADMMSlackvsOSPQ}{1}}(\texttt{ADMMSlack}: Differences with \OSQP)
},
coltitle=black, 
boxrule=2pt,
colframe=black!6, % 
enhanced, colback=black!2, boxsep=0pt, left=6pt, right=6pt]
The standard \OSQP~\cite{Stellato2020}  ADMM scheme  would introduce a copy $\tilde\xi$ of $\xi$, and use a primal  update that minimizes the Lagrangian over $(\tilde x, \tilde z, \tilde\xi)$, solving a larger linear system. Instead, our ADMM scheme   for handling slack variables is inspired from \cite{LewSlackADMM2025}. It is easier to implement as it only consists of a small change to the projection step. 
\end{tcolorbox}

\tcbset{
diffmpcOuterFull/.style={
    enhanced,
    coltitle=black,
    boxrule=3pt,
    colframe=gray!16,
    colback=gray!8,
    colbacktitle=gray!18,
    boxsep=0pt,
    left=6pt,
    right=6pt,
    top=2pt,
    bottom=2pt,
},
diffmpcForwardFull/.style={
    enhanced,
    coltitle=black,
    boxrule=3pt,
    colframe=ForestGreen!10,
    colback=ForestGreen!4,
    colbacktitle=ForestGreen!10,
    boxsep=0pt,
    left=6pt,
    right=6pt,
    top=2pt,
    bottom=2pt,
},
diffmpcADMMFull/.style={
    enhanced,
    coltitle=black,
    boxrule=3pt,
    colframe=ForestGreen!10,
    colback=ForestGreen!4,
    colbacktitle=ForestGreen!10,
    boxsep=0pt,
    left=6pt,
    right=6pt,
    top=2pt,
    bottom=2pt,
}
}
\definecolor{DiffMPCPurple}{HTML}{F2CCFF}
\definecolor{DiffMPCPurpleBorder}{HTML}{8200D8}

\tcbset{
diffmpcBackwardFull/.style={
    enhanced,
    coltitle=black,
    boxrule=3pt,
    colframe=DiffMPCPurpleBorder!12,
    colback=DiffMPCPurpleBorder!4,
    colbacktitle=DiffMPCPurpleBorder!10,
    boxsep=0pt,
    left=6pt,
    right=6pt,
    top=2pt,
    bottom=2pt,
}
}

\begin{figure}[!t]
\centering
\begin{tcolorbox}[
diffmpcOuterFull,
title={
\texttt{DiffMPC} (Differentiable MPC Solver)
\phantom{${}^1$}}
]

\begin{tcolorbox}[
diffmpcForwardFull,
title={
\textbf{Alg. \customlabel{alg:sqp}{1}}
\texttt{SQP} (Forward Pass)
\phantom{${}^1$}
}
]
\begin{algorithmic}[1]
\Statex \hspace{-5mm}\textbf{Inputs}: $\OCP$, tolerance $\epsilon$, max SQP iterations, linesearch parameters
\Statex \hspace{-5mm}\textbf{Initial guess}: primal-dual guess $(x,y)$

\While{not converged}
    \State \QP $\gets$ Convexify \OCP
    \newalgcomment{Sec.}{sec:solver:qp}

    \State $(x^\star,\xi^\star,y^\star)\gets$ Solve \QP via \ADMM
    \newalgcomment{Alg.}{alg:admm}

    \State $(x,y)\gets\text{Linesearch}(x^\star,y^\star,x,y)$

    \State Convergence Check
    \newalgcomment{Eq.}{eq:sqp_stop}
\EndWhile

\State \textbf{Return}: solution $(x,\xi,y)$
\end{algorithmic}
\end{tcolorbox}

\begin{tcolorbox}[
diffmpcBackwardFull,
title={
\textbf{Alg. \customlabel{alg:backward}{2}}
Active-Set \texttt{VJP} (Backward Pass)
\phantom{${}^1$}
}
]
\begin{algorithmic}[1]
\Statex \hspace{-5mm}\textbf{Inputs}: solution $(x,\xi,y)$, parameters $\theta$, upstream gradient $\partial\mathcal L/\partial x$

\State Identify active inequalities $\mathcal A$ 
\newalgcomment{Sec.}{subsec:backward}

\State Form the active-set KKT conditions $F(w,\theta)=0$

\State Assemble $\partial F/\partial w$ and $\partial F/\partial\theta$

\If{slack variables are disabled}
    \State Solve the no-slack linear system
    \newalgcomment{Eq.}{eq:bwd:noslack_kkt_matrix}
\Else
    \State Solve the slack linear system     \newalgcomment{Eq.}{eq:bwd:slack_reduced_system}
\EndIf

\State Compute the \texttt{VJP}
\newalgcomment{Eq.}{eq:loss_gradient}

\State \textbf{Return}: gradients $\partial\mathcal L/\partial\theta$
\end{algorithmic}
\end{tcolorbox}

\end{tcolorbox}
\vspace{-7mm}
\end{figure}
\begin{figure}[!t]
\centering
\vspace{5pt}
\begin{tcolorbox}[
diffmpcADMMFull,
title={
\textbf{Algorithm \customlabel{alg:admm}{3}}
\ADMM (\QP Solver)
\phantom{${}^1$}}
]
\begin{algorithmic}[1]
\Statex \hspace{-5mm}\textbf{Inputs}: $P,q,A,l,u,\sigma,\rho_f,\rho_g,\alpha,\gamma_\xi,\delta_\xi$
\Statex \hspace{-5mm}\textbf{Initial guess}: $x^0,z^0,y^0$
\Statex \hspace{-5mm}\textbf{Partition}: 
$z\,{=}\,\AverageSmallMatrix{z_f\\z_g}$, 
$y\,{=}\,\AverageSmallMatrix{y_f\\y_g}$, 
$A\,{=}\,\AverageSmallMatrix{C\\G}$, 
$l\,{=}\,\AverageSmallMatrix{c\\\underline G}$, 
$u\,{=}\,\AverageSmallMatrix{c\\\overline G}$

\While{not converged}
  \State \textbf{Primal update}: 
  	\Statex $\tilde x^{k+1}\gets \text{Solve}(S\tilde x=\eta^k)$
  \newalgcomment{Eq.}{eq:admm:schur}
  	\Statex $\tilde z^{k+1}=A\tilde x^{k+1}$

  \State \textbf{Slack update}: with 
  $\hat x^{k+1}=\alpha\tilde x^{k+1}+(1{-}\alpha)x^k$ and
  $\hat z^{k+1}=\alpha\tilde z^{k+1}+(1{-}\alpha)z^k$,
    \begin{align*}
    x^{k+1} &\leftarrow \hat x^{k+1}, 
    \qquad 
    z_f^{k+1}\leftarrow c,
    \\
    z_g^{k+1} &\leftarrow 
    \begin{cases}
    \Pi_{[\underline G,\overline G]}
    (\hat z_g^{k+1}+\rho_g^{-1}y_g^k),
    & \text{if }\delta_\xi=0,
    \\[1mm]
    \widetilde\Pi^{\gamma_\xi}_{[\underline G,\overline G]}
    (\hat z_g^{k+1}+\rho_g^{-1}y_g^k),
    & \text{if }\delta_\xi=1,
    \end{cases}
    \\
    \xi^{k+1} &\leftarrow 
    \begin{cases}
    0,
    & \text{if }\delta_\xi=0,
    \\[1mm]
    \Delta\widetilde\Pi^{\gamma_\xi}_{[\underline G,\overline G]}
    (\hat z_g^{k+1}+\rho_g^{-1}y_g^k),
    & \text{if }\delta_\xi=1.
    \end{cases}
    \end{align*}

  \State \textbf{Dual update}: 
  $y^{k+1}\leftarrow y^k+\rho\big(\hat z^{k+1}-z^{k+1}\big)$

  \State Update residuals, check convergence, and update $\rho$ when scheduled
  \newalgcomment{Eq.}{eq:admm:residuals}
\EndWhile

\State \textbf{Return}: solution $(x^{k+1},\xi^{k+1},y^{k+1})$
\end{algorithmic}
\end{tcolorbox}
\vspace{-7mm}
\end{figure}

\subsubsection{Primal update}
Minimizing the augmented Lagrangian $\mathcal L_{\sigma,\rho}$ over $(\tilde x,\tilde z)$ gives an
equality-constrained QP, whose KKT conditions give the linear system
\begin{equation}
\begin{bmatrix}
P+\sigma I & A^\top\\
A & -\rho^{-1}I
\end{bmatrix}
\begin{bmatrix}
\tilde x^{k+1}\\ \nu^{k+1}
\end{bmatrix}
=
\begin{bmatrix}
\sigma x^k-q\\
z^k-\rho^{-1}y^k
\end{bmatrix}.
\end{equation}
Thanks to the bottom-right block $-\rho^{-1}I$, we can eliminate the multipliers $\nu^{k+1}$, which gives the Schur-complement
system
\begin{equation}\label{eq:admm:schur}
S\tilde x^{k+1}=\eta^k,
\end{equation}
where the $k$ superscript denotes ADMM iterations and
\begin{align*}
S&=P+\sigma I+A^\top\rho A,
\ \ \ \ 
\eta^k=\sigma x^k-q+A^\top(\rho z^k-y^k).
\end{align*}
After solving \eqref{eq:admm:schur}, we recover
\begin{equation}
\tilde z^{k+1}=A\tilde x^{k+1}.
\end{equation}

The matrix $S$ is block-tridiagonal thanks to the sparsity of $(P,C,G)$, enabling efficient solutions to \eqref{eq:admm:schur} on the GPU (where $\Theta,\Phi$ are defined in Appendix~\ref{apdx:admm}):
\[
S =
\AverageSmallMatrix{
\Theta_0 & \Phi_0^\top
\\[-2mm]
\Phi_0 & \Theta_1 & \ddots
\\[-1mm]
& \ddots & \ddots & \Phi_{N-1}^\top
\\[-1mm]
& & \Phi_{N-1} & \Theta_N
}.
\]

\begin{tcolorbox}[
title={% 
\textbf{Remark \customlabel{remark:ImplicitDynamics}{2}}(Implicit dynamics and  KKT system sparsity)
},
coltitle=black, 
boxrule=2pt,
colframe=black!6, % 
enhanced, colback=black!2, boxsep=0pt, left=6pt, right=6pt]
Implicit integrators couple neighboring stages through both $x_t$, $u_t$ and $x_{t+1}$, $u_{t+1}$, which breaks the temporal decoupling that Riccati recursions require. However, the resulting $A_t^{-}$, $A_t^{+}$ Jacobian blocks preserve the block-bidiagonal structure of $C$ and the block-tridiagonal structure of $S$. Hence, implicit dynamics still enable GPU parallelism through the KKT primal solve even though they are incompatible with sequential Riccati-based methods.
\end{tcolorbox} 

\subsubsection{Slack update}
Given $\alpha\in(0,2)$, we first form the over-relaxed variables
\begin{equation}
\hat x^{k+1}
=
\alpha\tilde x^{k+1}+(1-\alpha)x^k,
\quad
\hat z^{k+1}
=
\alpha\tilde z^{k+1}+(1-\alpha)z^k .
\end{equation}

The slack update then consists of  minimizing $\mathcal L_{\sigma,\rho}$ over  $(x,z,\xi)$, which gives the following updates.  

First, minimizing $\mathcal L_{\sigma,\rho}$ over  $x$ gives
\begin{align}
x^{k+1}
&\leftarrow
\hat x^{k+1} + \sigma^{-1}w^k=\hat x^{k+1},
\end{align}
noting that $w\equiv 0$ (see Appendix~\ref{apdx:admm} for details).

Second, we split the equality and inequality constraints into two blocks
$$
z\,{=}\,\AverageSmallMatrix{z_f\\z_g}, 
y\,{=}\,\AverageSmallMatrix{y_f\\y_g}, 
A\,{=}\,\AverageSmallMatrix{C\\G}, 
l\,{=}\,\AverageSmallMatrix{c\\\underline G}, 
u\,{=}\,\AverageSmallMatrix{c\\\overline G}, 
\rho\,{=}\,\AverageSmallMatrix{
\rho_f I& 0
\\
0 & \rho_g I
},
$$
with the subscripts $f$ and $g$ denoting correspondence to the equality and inequality constraints, respectively.

For the equality block, minimizing $\mathcal L_{\sigma,\rho}$ over $z_f$ results in $z_f^{k+1}\gets c,$ and as such we do not store it explicitly.   
For the inequality block, when minimizing $\mathcal L_{\sigma,\rho}$ over $(z_g,\xi)$, we distinguish two cases:
\begin{itemize}
	\item % 
	Without slack variables ($\delta_\xi=0$),  we obtain
\begin{align}
\xi^{k+1}\leftarrow 0,
\quad 
z_g^{k+1} \leftarrow 
    \Pi_{[\underline G,\overline G]}
    (\hat z_g^{k+1}+\rho_g^{-1}y_g^k),
\end{align}
where $\Pi$ is the standard projection
\begin{align}
\Pi_{[\underline G,\overline G]}(\tilde{z})
=
\begin{cases}
\tilde{z}& \text{if }\underline G \leq \tilde{z} \leq \overline G,
\\
\underline G
& \text{if }\tilde{z}< \underline G,
\\
\overline G
& \text{if }\tilde{z}>  \overline G.
\end{cases}
\end{align}
	\item % 
	With slack variables ($\delta_\xi=1$),
by minimizing $\mathcal L_{\sigma,\rho}$ over  $(z_g,\xi)$, we get the smoothed projection \cite{LewSlackADMM2025}:
\begin{equation} \begin{split}
\xi^{k+1}&\leftarrow 
\Delta\widetilde\Pi^{\gamma_\xi}_{[\underline G,\overline G]}
    (\hat z_g^{k+1}+\rho_g^{-1}y_g^k),
\\ 
z_g^{k+1} &\leftarrow 
    \widetilde\Pi^{\gamma_\xi}_{[\underline G,\overline G]}
    (\hat z_g^{k+1}+\rho_g^{-1}y_g^k),
\end{split} \end{equation}
where
\begin{subequations}\label{eq:proj:tilde}
\begin{align}
\widetilde\Pi^{\gamma_\xi}_{[\underline G,\overline G]}(v)
&=
\frac{\rho_g}{\gamma_\xi+\rho_g}v
+
\frac{\gamma_\xi}{\gamma_\xi+\rho_g}
\Pi_{[\underline G,\overline G]}(v),
\\
\Delta\widetilde\Pi^{\gamma_\xi}_{[\underline G,\overline G]}(v)
&=
\frac{\rho_g}{\gamma_\xi+\rho_g}
\bigl(\Pi_{[\underline G,\overline G]}(v)-v\bigr).
\end{align}
\end{subequations}
Note that  $\gamma_\xi\to\infty$ recovers the hard projection.
\end{itemize} 

\begin{tcolorbox}[
title={
\textbf{Remark \customlabel{remark:SlacksLearning}{3}}(Slack variables and learning stability)
},
coltitle=black, 
boxrule=2pt,
colframe=black!6, % 
enhanced, colback=black!2, boxsep=0pt, left=6pt, right=6pt]
Slack variables are essential not only for solver robustness but for learning stability. Batched RL and BO rollouts can encounter infeasible states due to model mismatch and parameter exploration. A solver that crashes on infeasibility produces undefined gradients and halts training, motivating the use of slack variables for training and deployment. 
\end{tcolorbox} 

\subsubsection{Dual update}
It is as follows, where $w$ vanishes after the first iteration, so we only store the multipliers $y$:
\begin{subequations}\label{eq:admm:derivations:dual_updates_full}
\begin{align}
w^{k+1} &\leftarrow w^k + \sigma(\hat x^{k+1}-x^{k+1})=0,
\label{eq:admm:derivations:dual_w}
\\
y^{k+1} &\leftarrow y^k + \rho(\hat z^{k+1}-z^{k+1}).
\label{eq:admm:derivations:dual_nu}
\end{align}
\end{subequations}

\textbf{Step-size parameters selection}. 
The performance of ADMM highly depends on the choice of the step-size parameters $(\sigma,\rho)$. 
We found the parameter selection rule in \cite{Stellato2020} to work well in practice. Hence, we fix $\sigma=10^{-6}$ and use 
$$
\rho_f = 10^3\bar\rho,
\qquad
\rho_g=\bar\rho,
$$
with initial $\bar\rho=0.1$. 
We adapt $\bar\rho$ every fixed number  $k_{\rho}^{\textrm{iter}}\geq1$ of ADMM iterations to balance the primal and dual residuals. At each scheduled interval, we compute a candidate
\begin{equation}
\scalebox{1.0}{$
\label{eq:rho:update}
\bar\rho_{\textrm{new}}
=\Pi_{[\underline{\rho},\overline{\rho}]}\Bigg(
\bar\rho^{k}
\sqrt{
\frac{
r_{\mathrm{prim}} / \norminf{(Ax, z)}
}
{
r_{\mathrm{dual}}
/
\|
(Px,A^\top y,q,\gamma_\xi\xi,[0;\delta_\xi I]y)
\|_\infty
}
}\Bigg),
$}
\end{equation}
which we accept ($\bar\rho^{k+1}\gets\bar\rho_{\textrm{new}}$) if the primal and dual residuals have not yet satisfied the convergence criteria \eqref{eq:admm:termination:slack}, and if $\max\big(\frac{\bar\rho_{\textrm{new}}}{\bar\rho^{k}},\frac{\bar\rho^{k}}{\bar\rho_{\textrm{new}}}\big)$ is higher than a threshold value $k_{\rho}^{\textrm{ratio}}$. The last check measures how large the multiplicative change in $\bar\rho$ would be. The threshold values and the bounds in \eqref{eq:rho:update} are user and problem dependent. We use $\underline{\rho}=10^{-6}$ $\overline{\rho}=10^{6}$, $k_{\rho}^{\textrm{iter}}=25$, and $k_{\rho}^{\textrm{ratio}}=5$ by default. The ADMM literature shows similar strategies \cite{Stellato2020,Boyd_admm, Bravo2022}.  

\subsection{Linesearch and Convergence Check}\label{sec:solver:linesearch}
We use a standard backtracking linesearch as described in \cite[Algorithm 18.3]{Nocedal2006} and Appendix \ref{apdx:linesearch}. 
We terminate the SQP loop once the first-order optimality conditions are sufficiently satisfied, where $\|g(x)\|_{\infty,[\underline g,\overline g]}$ denotes the maximum componentwise violation of the inequality bounds, and $\epsilon_c>0$ is a user-defined tolerance:
\begin{subequations}
    \label{eq:sqp_stop}
\begin{gather}
\|\nabla\ell(x)+y_f^\top \nabla f(x) +y_g^\top\nabla g(x)\|_\infty\le\epsilon_c, 
\\ 
\|f(x)\|_\infty\le \epsilon_c,\ \  \|g(x)\|_{\infty,[\underline g,\overline g]}\le \epsilon_c,
\end{gather}
\end{subequations}
This convergence criterion is standard in the optimization literature~\cite{Nocedal2006,Jordana2025}.

\begin{tcolorbox}[
title={% 
\textbf{Remark \customlabel{remark:Rescaling}{4}}(Rescaling)
},
coltitle=black, 
boxrule=2pt,
colframe=black!6, % 
enhanced, colback=black!2, boxsep=0pt, left=6pt, right=6pt]
Problem rescaling by normalizing states and controls is a standard practice in aerospace trajectory optimization \cite{fanger2026augmented} that meaningfully improves solver convergence. For the racing \OCP, where state variables span several orders of magnitude, poor scaling inflates condition numbers and slows down convergence. % 
Our solver leverages rescaling internally to improve robustness, while keeping this step transparent and optional through the user-facing JAX API.
\end{tcolorbox} 

\subsection{Backward Pass: Computing Sensitivities}\label{subsec:backward}

We compute sensitivities of the solution $(x,\xi)$ of $\OCP$ with respect
to parameters $\theta$ via implicit differentation, based on the function theorem (IFT)~\cite{krantz2002implicit}. At convergence of SQP,
the solution satisfies the $\OCP$ KKT conditions. We differentiate these KKT conditions while keeping the
active set fixed, like~\cite{amos2018differentiable,frey2025differentiable}.

\textbf{Active-set KKT system.}
Let $\mathcal A$ denote the set of active inequality constraints at the converged
solution. 
From identifying the active inequality constraints $g_{\mathcal A}$, as described in Appendix~\ref{apdx:backward}, we rewrite these inequality constraints as equality
constraints:
\begin{align*}
g_{\mathcal A}(x)&=0
&&\text{if }\delta_\xi=0 \text{ (without slack variables)},
\\
g_{\mathcal A}(x)+\Sigma_{\mathcal A}\xi_{\mathcal A}&=0
&&\text{if }\delta_\xi=1 \text{ (with slack variables)}, 
\end{align*}
where $\Sigma_{\mathcal A}$ is diagonal with entries $+1$ for lower-bound
activations and $-1$ for upper-bound activations. 
The reduced active-set KKT conditions of \OCP are 
\begin{subequations}
\label{eq:bwd:kkt}
\begin{align}
\nabla_x L(x,y_f,y_{\mathcal A}) &=0, \label{eq:bwd:kkt_stationarity}
\\
f(x)&=0,
\label{eq:bwd:kkt_eq}
\\
g_{\mathcal A}(x)-\delta_\xi\gamma_\xi^{-1}y_{\mathcal A}&=0,
\label{eq:bwd:kkt_active}
\end{align}
\end{subequations}
where $L$ is the Lagrangian
$L(x,y_f,y_{\mathcal A}) = \ell(x)+y_f^\top f(x)+y_{\mathcal A}^\top g_{\mathcal A}(x).$
We eliminate the slack stationarity
condition $\gamma_\xi \xi_{\mathcal A}
+
\Sigma_{\mathcal A}y_{\mathcal A}=0$ in the presence of slack variables.   
We then write \eqref{eq:bwd:kkt} compactly as
\begin{equation}
F(w,\theta)=0,
\qquad
w=(x,y_f,y_{\mathcal A}).
\end{equation}

\textbf{Gradients computation}: 
We can now use standard techniques to obtain gradients, see e.g. \cite{adabag2025differentiable,amos2018differentiable}. 
By the chain rule, differentiating $F(w(\theta),\theta)=0$ with respect to $\theta$ gives
\begin{equation}
\frac{\partial F}{\partial w}
\frac{\partial w}{\partial \theta}
+
\frac{\partial F}{\partial \theta}
=0
\implies
\label{eq:bwd:ift}
\frac{\partial w}{\partial \theta}
=
-
\left[
\frac{\partial F}{\partial w}
\right]^{-1}
\frac{\partial F}{\partial \theta}.
\end{equation}
 The above computation corresponds to a Jacobian-Vector Product (JVP), and requires solving a large linear system. Instead, the gradient of a loss $\nabla_\theta \mathcal L(x_\theta,\xi_\theta)$ can be more efficiently computed via a Vector-Jacobian Product (VJP):
\begin{align}\label{eq:loss_gradient}
\hspace{-2mm}
\nabla_\theta \mathcal L^\top 
&= 
-\frac{\partial F}{\partial \theta}^\top
\left(
\left[\frac{\partial F}{\partial w} \right]^{-1}
\AverageSmallMatrix{
\nabla_x\mathcal L\\
\nabla_{\xi_{\mathcal A}}\mathcal L\\
0
}
\right)
= 
-\frac{\partial F}{\partial \theta}^\top
\AverageSmallMatrix{
\eta_x\\
\eta_\xi\\ 
\eta_y
},
\end{align}
which also follows from the chain rule, and requires solving the following where $\eta_y=(\eta_{y_f},\eta_{y_{\mathcal A}})$, $H = \nabla_{xx}^2 L(x,y_f,y_{\mathcal A})$: 
\begin{equation}
\label{eq:bwd:LS}
\AverageSmallMatrix{
H & \nabla_x f^\top & \nabla_x g_{\mathcal A}^\top\\
\nabla_x f & 0 & 0\\
\nabla_x g_{\mathcal A} & 0 & -\delta_\xi\gamma_\xi^{-1}I
}
\begin{bmatrix}
\eta_x\\
\eta_\xi\\ 
\eta_y
\end{bmatrix}
=
\begin{bmatrix}
\nabla_x\mathcal L\\
\nabla_{\xi_{\mathcal A}}\mathcal L\\
0 
\end{bmatrix}.
\end{equation}

\textbf{Linear systems in the forward  and backward passes}.
Note the difference in sparsity structure:
\vspace{-3mm}
\begin{multicols}{2}
\begin{center}
\textit{Primal Linear System\\(Forward Pass)}
\end{center}
\[
\begin{bmatrix}
P+\sigma I & A^\top\\
A & -\rho^{-1}I
\end{bmatrix}
\]

\begin{center}
\textit{Adjoint Linear System\\(Backward Pass)}
\end{center}
\[
\begin{bmatrix}
H & \nabla f^T\\
\nabla f & 0
\end{bmatrix}
\]
\end{multicols}
\noindent For the forward pass, the block  $\rho^{-1}I$ is amenable to a Schur reduction with a block-tridiagonal structure. Unfortunately, that block is zero for the backward pass. Eliminating the top left block instead generally destroys block-tridiagonality whenever $H$ contains cross-time coupling from implicit dynamics or control-rate costs. Thus, PCG primitives designed for the forward system~\cite{adabag2024mpcgpu,du2026gato,adabag2025differentiable} do not directly apply to our backward pass, motivating the use of \texttt{cuDSS}.% 

\begin{tcolorbox}[
title={% 
\textbf{Remark \customlabel{remark:co-designFwdBwd}{5}}(Co-design of forward and backward passes)
},
coltitle=black, 
boxrule=2pt,
colframe=black!6, % 
enhanced, colback=black!2, boxsep=0pt, left=6pt, right=6pt]
The incompatibility of PCG with the backward pass implies that the forward and backward passes must be co-designed from the start. Choosing \texttt{cuDSS} as a uniform solver, serving both the forward primal update and the adjoint linear system, is a direct consequence. 
\end{tcolorbox}

\subsection{Algorithm-Implementation-Learning Co-Design}\label{sec:solver:design}
The development of \solverName~is a three-way co-design of the optimization algorithm, the GPU implementation, and the learning pipeline. Unlike hardware/control co-design~\cite{Bravo2024, Bjelonic2023, DeVincenti2021}, where mechanical and control tasks have clear physical boundaries, the boundary between solver and differentiable policy here is subtle: the same ADMM splitting that gives the forward pass its block-tridiagonal structure determines whether the backward pass is tractable, and the same cuDSS factorization that solves the primal update also solves the adjoint system. Designing these concurrently, rather than building a fast solver and then asking how to differentiate it, is what allows \solverName~to simultaneously achieve GPU acceleration, constraint expressiveness, and differentiability.

The architecture of \solverName~is summarized in Figure~\ref{fig:hero}. The JAX layer hosts the outer SQP loop, problem-data linearization via autodiff of $(\ell, f, g, h)$, and \texttt{vmap}-based batching across problem instances. A  C++ FFI (Foreign Function Interface) bridge passes JAX-managed device buffers to the CUDA implementation. The CUDA layer hosts the entire ADMM inner loop as a fused implementation, with custom parallel-over-time kernels for matrix-vector products with $P$, $C$, $G$, $A$, $A^\top$, $S$ that exploit the time-induced block sparsity established in Section~\ref{sec:solver:qp}. The same compiled artifact supports online deployment (batch one, low latency) and batched offline evaluation (batch $B$, throughput proportional to GPU occupancy) from a single API.

Two important co-design choices of \solverName~are:

\textbf{Linear-system primitive.} The forward primal update \eqref{eq:admm:schur} and the backward linear system \eqref{eq:bwd:LS} both require sparse linear-system solves, but as established in Section~\ref{subsec:backward} the backward system has a structure that rules out the PCG-based primitives prior GPU MPC work relies on~\cite{adabag2024mpcgpu,adabag2025differentiable}. We use direct sparse factorization via \texttt{cuDSS}~\cite{cudss} as a uniform primitive across both passes, called from within our fused CUDA ADMM loop through \texttt{cuDSS}'s host API. This approach trades a constant-factor solve-time penalty for three benefits: (1) generality across the explicit-vs-implicit and decoupled-vs-coupled-cost axes, (2) robustness near constraint activation where iterative methods can exhibit long-tail convergence, and (3) batched factorization across  many problem instances. \texttt{cuDSS}'s symbolic factorization is reused across ADMM iterations and refactorized only when $\rho$ adapts.

\textbf{Fused CUDA ADMM Loop.} An earlier version hosted ADMM in JAX, dispatching only the \texttt{cuDSS} solves to CUDA via FFI. Per-iteration kernel-launch overhead dominated solve time. We fused the entire ADMM loop into a single CUDA implementation that calls \texttt{cuDSS} through its host API and reuses the symbolic factorization across iterations, yielding a $4-8\times$ speedup at batch sizes $256-512$ relative to the JAX-FFI variant while preserving the user-facing JAX API.

Overall, we jointly select the ADMM splitting, the slack-variable smoothed projection of~\cite{LewSlackADMM2025}, and the implicit-differentiation-based backward pass to admit time-induced block sparsity, with a forward-pass linear system whose Schur reduction preserves block-tridiagonality even with implicit dynamics or cross-time costs, both with underlying \texttt{CUDA} for performance and a \texttt{JAX} frontend for easy integration with learning-based tools.

\subsection{Reinforcement and Imitation Learning}\label{subsec:rl_il}

\solverName{} can be used as a \textit{differentiable policy class} for  reinforcement learning (RL) and imitation learning (IL):
\begin{align*}
&\textbf{RL:}&& \max_\theta\ \mathbb{E}\Big[\sum_{t=1}^H R(x_t,\pi_0^\theta(x_t))\Big],\ x_{t+1}{=}\texttt{Sim}(x_t,\pi^\theta(x_t)),\\
&\textbf{IL:}&& \min_\theta\ \mathbb{E}\Big[\|(\hat u_0,\dots,\hat u_T)-\pi_{0:T}^\theta(x_0)\|^2\Big],
\end{align*}
where $R$ is the reward function, $\pi_{0:T}^\theta(x_0)$ is the $\OCP$ control-input solution, \texttt{Sim} is a differentiable simulator, and $(\hat u_0,\dots,\hat u_T)$ is expert demonstration data. Compared to black-box neural networks, this MPC policy class leverages optimal-control structure for inductive bias and zero-shot generalization across problem instances. Our GPU-accelerated solver supports the larger batch sizes and expressive models these data-driven workflows might require.

\section{Simulation Experiments}\label{sec:sim}

We first validate \solverName{} in simulation using a constrained drone task, a linear-system RL scaling benchmark, a humanoid imitation-learning task, and a neural-network (NN) training task. % 
Timings are collected using an NVIDIA GeForce RTX 5090 GPU and a 12th Gen Intel(R) Core(TM) i9-12900K CPU with Ubuntu 22.04 and CUDA 13.0. % 

\subsection{Constrained Drone Obstacle Avoidance}
\begin{figure}[!t]
\centering
\includegraphics[width=1.0\columnwidth]{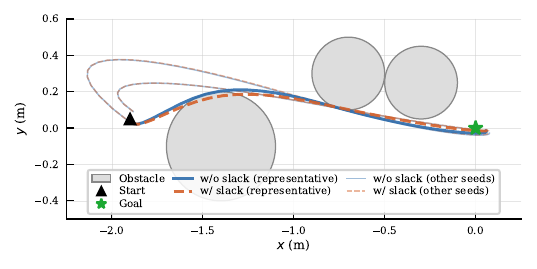}
\vspace{-30pt}
\caption{Drone obstacle avoidance: Closed-loop trajectories from random initial conditions (i.e., seeds). }% 
\label{fig:drone}
\vspace{-10pt}
\end{figure}

We validate that \solverName~handles state inequalities, control constraints, and slack variables on a 6-DoF drone obstacle-avoidance task with three circular obstacles in $(x,y)$, initial and final state constraints, and control box constraints. Figure~\ref{fig:drone} shows representative trajectories from running closed-loop simulations from noisy initial states using a straight-line warm start and RK4 integration. Adding slack variables can increase the rate of solver convergence at the cost of increased constraint violation, as the relaxed constraints allow the trajectory to trade feasibility for convergence. We provide a \texttt{Jupyter} notebook that replicates this example and serves as an introductory tutorial.

\subsection{Linear-System RL Scaling}\label{subsubsec:exp:linear}

We next compare \solverName{}  against state-of-the-art differentiable MPC baselines: the PyTorch-based iLQR solver \texttt{mpc.pytorch}~\cite{amos2018differentiable}\footnote{\texttt{mpc.pytorch} excludes comparing on problems with state inequalities, implicit dynamics, or rate costs as it only supports control box constraints.} and the C-based solver \texttt{acados}~\cite{frey2025differentiable}. To enable a comparison against these solvers within their supported problem class, we benchmark on an RL task (Section \ref{subsec:rl_il}) including linear-quadratic optimal control problems with control input box inequality constraints. We do not compare against \texttt{trajax}~\cite{frostig2021trajax} or  \texttt{DiffMPC}~\cite{adabag2025differentiable} since these differentiable solvers do not support inequality constraints. The linear quadratic setting represents the common subclass that all the baseline solvers support, enabling a direct comparison of raw solver scaling without confounding factors from nonlinear dynamics such as Hessian approximation quality and linesearch behaviour.  While this task is more restrictive than the full feature set \solverName~supports, this evaluation provides a fair comparison on this easy subclass, with more challenging problem instances in the next sections.

The linear system's discretized dynamics are $x_{t+1} = A x_t + B u_t + b$ for $t = 0,\dots,T-1$, with $x_t \in \mathbb{R}^{n_x}$, $u_t \in \mathbb{R}^{n_u}$, and control bounds $\|u_t\|_\infty \leq u_{\max}$, with $u_{\max}\in\{1, 10\}$. The dynamics matrices are randomized as $A = I + 0.1\,\Delta A$, $\operatorname{vec}(\Delta A) \sim \mathcal{N}(0,I)$, with eigenvalues clipped to $|\lambda(A)| \in (0, 0.99]$; $\operatorname{vec}(B) \sim \mathcal{N}(0,I)$; $b \sim \mathcal{N}(0, 10^{-4} I)$; and initial conditions $x_0 \sim \mathcal{N}(0, 25 I)$. The stage cost is  $\ell_t(x,u) = x^\top Q x + \|u\|_2^2$ with $Q = \operatorname{diag}(\theta)$. The RL reward is $R(x,u) = -(\|x\|_2^2 + \|u\|_2^2)$. The learned parameters are $\theta = \operatorname{diag}(Q)$.

\begin{figure*}[t]
\centerline{\includegraphics[width=0.95\linewidth]{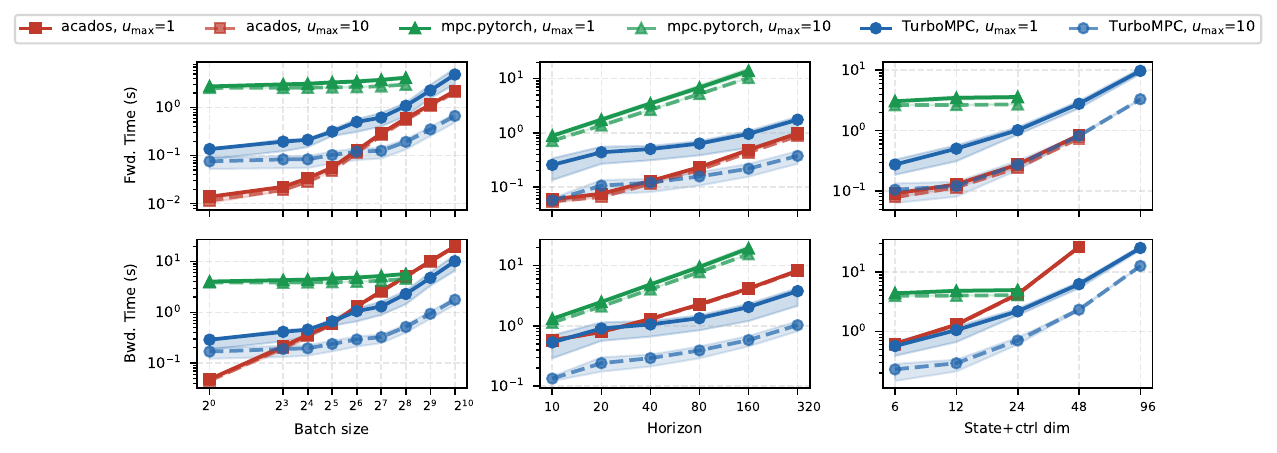}}
\vspace{-10pt}
\caption{Linear-system RL: mean solve time vs.\ batch size (left), planning horizon (center), state $+$ control dimension (right) for \solverName~(blue), \texttt{acados} (red), and \texttt{mpc.pytorch} (green) at $u_{\max}\in\{1,10\}$ with $\texttt{tol}=10^{-5}$.}
\label{fig:bench:ls-rl}
\vspace{-10pt}
\end{figure*}

\begin{figure*}[htbp]
\centering
\includegraphics[width=0.95\linewidth]{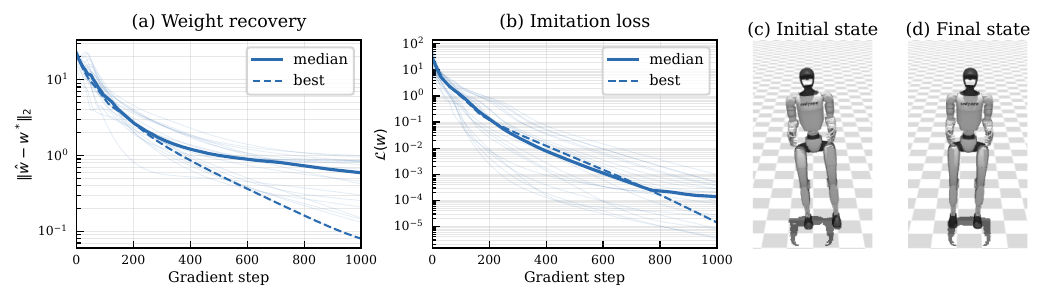}
\vspace{-10pt}
\caption{Humanoid SRB IL: (a) weight recovery and (b) imitation loss over gradient steps for \solverName~across 20 initializations. Closed-loop balancing in MuJoCo with recovered weights: (c) initial perturbed state and (d) reached equilibrium state.}
\label{fig:il}
\vspace{-10pt}
\end{figure*}

The evaluation tackles the RL problem in Section \ref{subsec:rl_il}. Timing results measure the evaluation of closed-loop rollouts and reward gradient evaluation through the closed-loop MPC rollouts. 
The nominal configuration uses: batch size $B=64$, planning horizon $N=40$, episode length $H = 50$, and state-control dimensions $(n_x, n_u) =(8,4)$. Then, we sweep $B\in\{1,8,16,32,64,128,256,512,1024\}$, $N\in\{10,20,40,80,160,320\}$, $(n_x{+}n_u)\in\{6,12,24,48,96\}$ and the solver tolerance $\texttt{tol}\in\{10^{-3},10^{-5}\}$, with statistics computed over ten seeds per configuration. The solver tolerance of \solverName~corresponds to $\epsilon_c$~in \eqref{eq:sqp_stop} and the maximum number of ADMM iterations is set to $10^3$. Figure~\ref{fig:bench:ls-rl} compiles the timing results for each solver with solver tolerance $\texttt{tol}=10^{-5}$. This threshold produces both accurate gradients (see cosine similarity plot in Appendix~\ref{appendix:linear}) and fast solve times across both \solverName~and baselines. In contrast, lower tolerances like $\texttt{tol}=10^{-3}$ can be used for deployed applications of \solverName~(e.g., Fig.~\ref{fig:bench:ls-rl-1e-3}), where higher speed is desired and accurate gradients are not required.

\solverName's advantage grows with planning horizon, problem size, and batch size. Compared to \texttt{mpc.pytorch} the gain reaches $14.6\times$ and $9.4\times$ speedups for the forward and backward passes for $(B,N,n_x,n_u)=(64,160,8,4)$ with $u_{\max}=1.0$. On the other hand, \texttt{acados}, which exploits efficient CPU linear algebra and parallelization across cores, is the fastest for single-instance forward solves overall, although \solverName~catches up as the batch size, horizon, and problem size increase. For the batch-size and horizon sweeps, 
for $u_{\max}\,{=}\,1.0$, \solverName~is $2.0\times$ and $2.2\times$ faster for the backward pass at batch size $1024$ and horizon $320$, respectively. The speedup is even more significant for $u_{\max}\,{=}\,10.0$. 
On the largest configurations, the baselines fail: \texttt{mpc.pytorch} hits GPU memory limits, and \texttt{acados} cannot compile above $(n_x{+}n_u)=48$. \solverName~solves all sizes by leveraging GPU parallelism, time-induced sparsity, and \texttt{cuDSS} for the linear system solves. With a $10^{-3}$ tolerance, the results show similar trends, with \solverName~further outperforming baselines, achieving up to $15\times$ and $58\times$ speedups over \texttt{acados} and \texttt{mpc.pytorch}, respectively, as detailed in Appendix~\ref{appendix:linear}.   

\subsection{Differentiability on Constrained Nonlinear Systems}\label{subsubsec:exp:diff_nonlinear}
We demonstrate end-to-end differentiability on constrained nonlinear systems via a humanoid imitation-learning (IL) task and a neural-network (NN) training task.

\textbf{Humanoid Imitation Learning}: 
We recover expert cost weights $\theta^\star \in \mathbb{R}^{18}$ from  demonstrations of a Unitree G1 humanoid performing standing balance. Both expert and student MPCs use Single Rigid Body (SRB) centroidal dynamics with state $x \in \R^{12}$, and control $u \in \R^6$ (see Appendix~\ref{appendix:humanoid}). We collect $B=100$ expert demonstrations $(x_0^{(i)},u^{(i)})$ and minimize the imitation loss $\mathcal{L}(\theta) = \sum_{i=1}^B \frac{1}{N} \sum_{t=0}^{N-1} \|u_t(\theta; x_0^{(i)}) - u_t^{(i)}\|_2^2$ with respect to $\theta$ using  Adam  from $N_{\mathrm{inits}}=20$ random initializations ($\theta_0 \sim \mathcal{U}(0.5,12.0)$, applied in log-space).

Figure~\ref{fig:il} summarizes the results. At step 1000, the median weight-recovery error is $\|\hat{\theta}-\theta^\star\|_2=0.59$, with 19/20 runs below 1.0 and the best at 0.080. The median imitation loss reaches $1.4\times 10^{-4}$, with the best run at $3.4\times 10^{-6}$. Closed-loop balancing in MuJoCo XLA~\cite{todorov2012mujoco} from a perturbed initial state to equilibrium (Fig.~\ref{fig:il}, (c) (d)) demonstrates the recovered MPC weights produce a usable controller.
Differentiating through the constrained MPC embeds constraint-awareness into the recovered cost weights, capturing expert balancing behavior near the friction-cone boundary.

% The differentiability handling generalizes beyond IL, as demonstrated in the following example training neural networks. 
% % 
% The forward pass identifies the active set at convergence and yields constraint masks that the backward pass applies to produce sensitivities under active-set conditions. 
% % 
% 

\begin{figure}[t]
\centering
{% 
\includegraphics[width=0.53\columnwidth]{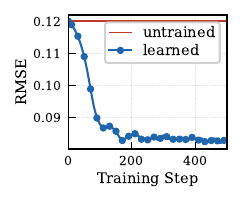}% 
\label{fig:pointmass:curve}}\hfil
{% 
\includegraphics[width=0.45\columnwidth]{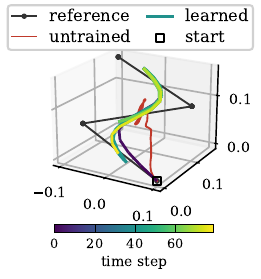}% 
\label{fig:pointmass:traj}}
\vspace{-4pt}
\caption{RL with a trainable MPC policy using a neural-network cost function on a 3D point-mass  system tracking a zig-zag reference. (Left) Two-cycle eval RMSE over $500$ Adam steps. (Right) Closed-loop trajectory after training: learned policy (blue) vs.\
untrained initialisation (grey).}
\label{fig:pointmass}
\vspace{-10pt}
\end{figure}

\textbf{Neural Network Training}: 
We train a \solverName{} policy using a neural-network cost function for a 3D point-mass system tracking a periodic zig-zag reference under input bounds
$\|u\|_\infty\!\le\!5$. The parameters $\theta$ are the weights of a
two-layer neural network that maps the state to the MPC's
state-tracking and control-effort cost weights,
trained by analytic policy gradient with closed-loop gradients flowing through \solverName{} to maximize a reference tracking reward.
With $B=4$ rollouts per Adam step at horizon $N=8$, the policy reduces the two-cycle closed-loop tracking RMSE from $1.2\!\times\!10^{-1}$\,m at initialization to $8.3\!\times\!10^{-2}$\,m after $500$ gradient steps, with control inputs satisfying constraints throughout. Figure~\ref{fig:pointmass} shows the learning curve and the converged closed-loop trajectory. Appendix~\ref{app:pointmass-rl} includes additional details. 

\section{Hardware Deployment: Autonomous Racing}\label{sec:racing}
Finally, we evaluate \solverName~on a 2019 Lexus LC 500 vehicle racing along the oval track in Figure~\ref{fig:racing:solvetimes:spinoutosqp}. This problem involves the full feature set covered by \OCP: state and control constraints, cross-time costs on control rates for smoothness, slack variables for recursive feasibility under disturbance, and implicit dynamics constraints to handle stiff dynamics.
First, we use \solverName{} to automatically tune the racing controller for faster driving performance via Bayesian optimization. Then, we evaluate the solver's scalability for long horizons. 

The vehicle is equipped with a GeForce RTX 4070 Ti Super GPU and an Intel Xeon E2278GE CPU @3.30GHz with a computer running Ubuntu, with state estimates provided by an OXTS GPS system. The powertrain, drivetrain, and suspension are unchanged from their production configuration. 
Experiments are performed on a closed-course.

We use a standard single-track vehicle model in curvilinear coordinates \cite{Goh2019,Dallas2023,Kobayashi2024,LewGreiffICRA2025}. 
The state and control inputs are
\begin{align*}
x&=
(
r, 
v,
\beta,
\omega_r,
\Delta F_z,
e,
\Delta\varphi,
t
),
\ \ 
u=
(
\delta, 
\tau_\textrm{rear},
\tau_\textrm{front} 
),
\end{align*}
parameterized over the track length with a node $(x_t,u_t)$ every three meters, where $r$ is the yaw rate, $v$ is the total velocity, $\beta$ is the sideslip, $\omega_r$ is the rear wheelspeed, $\Delta F_z$ is the load transferred from the front to rear axle, $e$ and $\Delta\varphi$ are the lateral and angle deviations to the reference, $t$ is time, $\delta$ is the steering angle, $\tau_\textrm{rear}$ is the total rear axle torque combining engine and braking torques, and $\tau_\textrm{front}$ is the rear brake torque. 
We model tire forces using a Fiala tire model~\cite{fiala1954seitenkraften}. 
The racing \OCP is the minimum-time problem
\begin{align}
\min_{x,u}
\ 
&\ell_T(x_N)+ 
\sum_{t=0}^{N-1}
\ell(x_t,u_t,u_{t+1}) + 
\sum_{t=1}^{N}
\gamma\|\xi\|^2
&&
\hspace{4mm} % 
\\[-1mm]
\text{s.t.}
\ \, 
&f(x_{t+1},x_t,u_t,u_{t+1},\mu)=0, &&\hspace{-8mm}
t = 0,\dots, N-1, \notag
\\
&(x_0,u_0)=(x_{\text{init}},u_{\text{init}}),  \notag
\\
&(x_{\textrm{min}}, u_{\textrm{min}})\leq (x_t,u_t)+\xi_t\leq (x_{\textrm{max}}, u_{\textrm{max}}),
&&\hspace{-2mm}t=1,\dots,N,  \notag
\end{align}
with  the costs
\begin{align}
\label{eq:cost}
\ell_T(x)&=\alpha\, t,
\ \,  \ell(x,u,u^+)
=(x-x_{\textrm{ref}})^\top Q(x-x_{\textrm{ref}})
\,+
\hspace{-2mm}
\\
&\hspace{-7mm} 
(u-u_{\textrm{ref}})^\top R(u-u_{\textrm{ref}})
\,+
(u-u^+)^\top W(u-u^+),
\nonumber
\end{align}
where $(Q,R,W)$ are diagonal cost matrices, $(x_{\textrm{ref}},u_{\textrm{ref}})$ is a state-control reference, and $\alpha$ is a terminal-time penalty. 
The racing \textbf{OCP} is solved recursively from the current $(x_{\text{init}},u_{\text{init}})$ and the plan $(u_0,\dots,u_N)$ is sent to the vehicle. A low-level controller executes the control $u$ at the current position along the track via linear interpolation of the latest MPC solution.  

This racing problem pushes the vehicle to its handling limits. As such, closed-loop performance is highly sensitive to the parameters $\theta = ($the tire friction coefficients $(\mu_{\textrm{front}},\mu_{\textrm{rear}})$, the terminal-time penalty $\alpha,$ and the diagonal entries of the state cost $Q$, control cost $R$), motivating the GPU-accelerated Bayesian optimization tuning of Section~\ref{sec:racing:autotuning}. We then stress-test the solver for long planning horizons to demonstrate the scalability of \solverName{} in Section~\ref{sec:racing:scaling}.

\subsection{Auto-tuning of the MPC parameters}\label{sec:racing:autotuning}

To avoid the slow manual tuning of $\theta = (\mu, \alpha, Q, R)$, we apply Bayesian optimization over GPU-batched closed-loop simulations leveraging \texttt{Optuna}'s~\cite{optuna_2019} Tree-structured Parzen Estimator (TPE) sampler~\cite{Bergstra2011}. The control-rate (or smoothness) cost weights $W$ % 
remain constant to smooth inputs.

All weights are optimized in log-space over their respective search intervals, with control weights for the torque channels tied together to reduce search dimensionality. At each iteration, a batch of $B = 8$ candidate parameter vectors is evaluated in parallel, with each candidate assessed over $N_{\mathrm{seed}} = 4$ closed-loop simulations with randomized tire friction coefficients $\mu_{\mathrm{sim}} \in [0.7, 1.4]$, initial track positions $s_0 \in [95, 135]$\,m, and initial velocities $v_0 \in [11, 20]$\,m/s, yielding $B \times N_{\mathrm{seed}} = 32$ parallel rollouts per batch. The simulation environments are sampled once at start and held fixed throughout, ensuring all candidates are evaluated under identical conditions.

\begin{table*}[t]
\centering
\caption{Baseline vs.\ trained MPC policy parameters.}
\label{tab:mpc_params}
\resizebox{\linewidth}{!}{% 
\begin{tabular}{lcccccccccccccc}
\toprule
 & $Q_r$ & $Q_v$ & $Q_\beta$ & $Q_{\omega_r}$ & $Q_e$ & $Q_{\Delta\varphi}$
 & $\alpha$
 & $R_\delta$ & $R_{\tau_{\textrm{rear}}}$  & $R_{\tau_{\textrm{front}}}$
 & $\mu_{\textrm{front}}$ & $\mu_{\textrm{rear}}$ \\
\midrule
Baseline
  & $10.0$   & $0.01$    & $1.0$    & $0.01$   & $10.0$  & $0.01$
  & $10.0$
  & $0.1$      & $1\times10^{-4}$ & $1\times10^{-5}$
  & $1.02$ & $1.08$ \\
Trained
  & $0.080$  & $0.016$   & $11.70$  & $0.449$  & $14.93$ & $13.72$
  & $129.30$
  & $0.018$    & $4.8\times10^{-7}$ & $4.8\times10^{-7}$
  & $1.081$ & $1.032$ \\
\midrule
Ratio
  & $\times 0.008$ & $\times 1.6$ & $\times 11.7$ & $\times 44.9$ & $\times 1.5$ & $\times 1372$
  & $\times 12.9$
  & $\times 0.18$ & $\times 0.005$ & $\times 0.048$
  & $+6\%$ & $-4\%$ \\
\bottomrule
\end{tabular}}
\end{table*}

The objective maximized by TPE is $\mathcal{R}(\theta) = \bar{R}(\theta) - k \cdot \sigma_R(\theta)$,  
where $\bar{R}(\theta)$ and $\sigma_R(\theta)$ are the mean and standard deviation of the per-seed rewards $\{R_j(\theta)\}_{j=1}^{N_{\mathrm{seed}}}$, and $k = 0.5$ balances average performance against consistency across conditions. Each per-seed reward is as follows, where $T_j \leq T_{\max}$ is the number of steps completed before a spin-out or off-track termination in seed $j$, $T_{\max}$ is the full rollout budget, and $\lambda = 0.9$ penalizes each missed step uniformly:
\begin{equation}
    R_j(\theta) = \sum_{t=0}^{T_j - 1} r_t - \lambda \max(0,\, T_{\max} - T_j).
\end{equation}

The per-step reward is as follows, where $e_t$ and $e_{\mathrm{ref}}$ are the actual and reference lateral deviations, respectively, $\delta_e = \min(e_{\mathrm{ref}} - e_{\min},\, e_{\max} - e_{\mathrm{ref}})$ is the half-width of the track boundary margin at the reference position, $\beta_t$ is the vehicle slip angle, and $v_t$ is the longitudinal velocity:
\begin{align}\label{eq:reward_step}
    r_t &= \alpha_{e,\beta} 
    \big(\hspace{-2pt}
    \min\!\left(\delta_e \,{-}\, |e_t {-} e_{\mathrm{ref}}|, 0\right)  + \min\!\left(\overline\beta \,{-}\, |\beta_t|, 0\right)
    \hspace{-2pt}\big)
    \,{+}\, \alpha_v\, v_t. \nonumber
\end{align}
The first two terms penalize leaving the track and reaching sideslip values above $\overline\beta = 0.55~\rm{rad}$, while the third term rewards higher speed. We use $(\alpha_{e,\beta}, \alpha_v) = (0.8, 0.14)$.% 

The entire rollout pipeline runs on the GPU, with dynamics integration via \texttt{Diffrax}~\cite{Kidger2022}, and the MPC solve and reward accumulation batched together inside a single \texttt{jax.lax.scan} and \texttt{jax.vmap} call across $B \times N_{\mathrm{seed}}$ instances. This GPU parallelism makes Bayesian optimization tractable with more than $256$ trials. 

\textbf{GPU acceleration enables faster hyperparameter search}. Using the default batch size $B=8$ on the GPU, the trial throughput is of ${\approx}150~\rm{trials/hour}$ or $24~\rm{secs./trial}$, while on the CPU the trial rate is of ${\approx}21~\rm{trials/hour}$ or $171~\rm{secs./trial}$. The GPU-over-CPU speedup is of ${\approx}7.12\times$, which represents a per-trial time reduction of ${\approx}86~\rm{secs.}$ The CPU runs used a PCG-based \texttt{JAX} solver backend (i.e., \texttt{admm\_jax\_loop\_pcg} backend for both the forward and backward passes) that we make available with our implementation, see Table~\ref{tab:backends}. 
In addition, we verified the GPU scalability by running the \texttt{Optuna} trainings with $B=1024$ batches at a rate of ${\approx}1,029~\rm{trials/hour}$ or $3.5~\rm{secs./trial}$ 
, which represents trial-throughput speedups of ${\approx}6.9\times$ and ${\approx}50\times$ over the 8-batch trainings on the GPU and CPU, respectively.

\begin{figure}[!t]
\centering
\includegraphics[width=1.0\columnwidth]{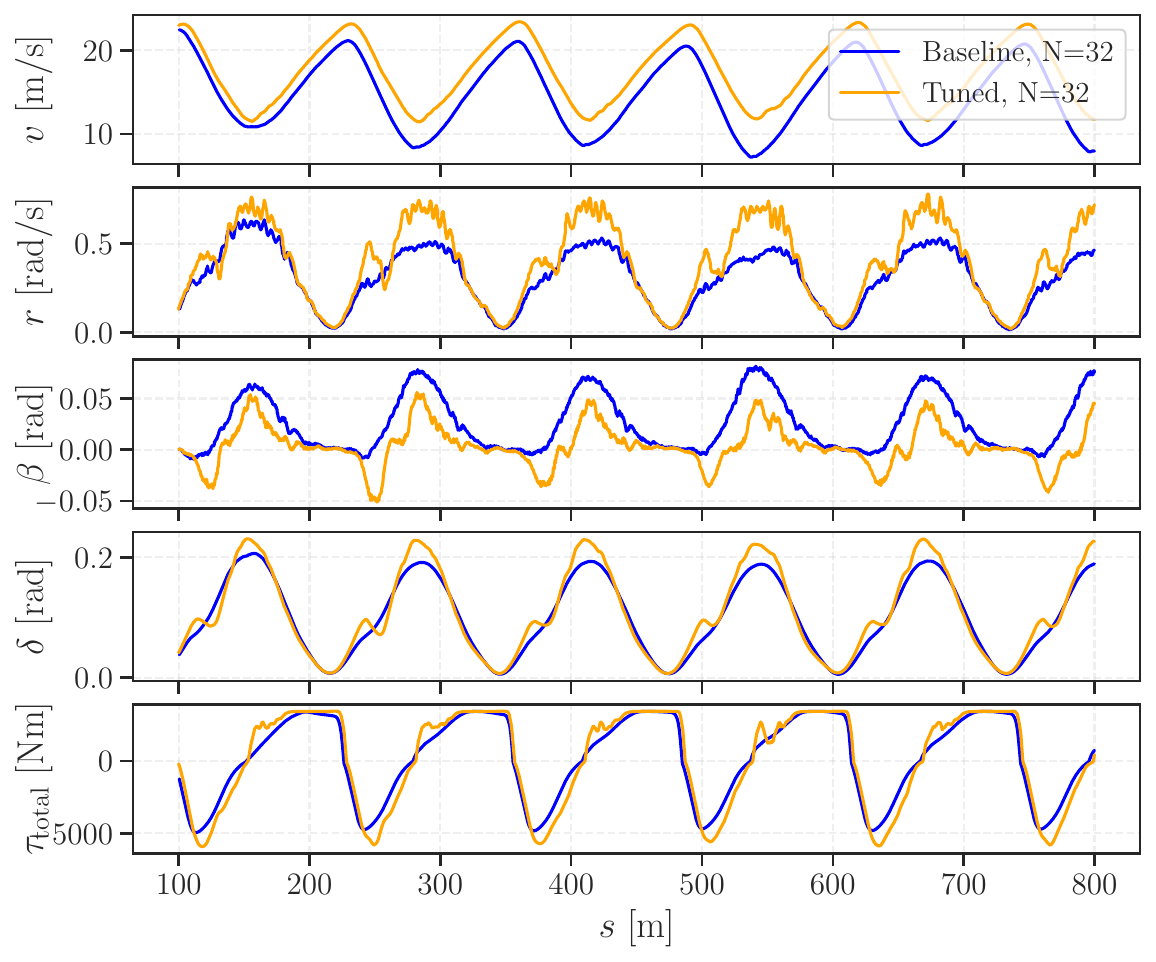}
\vspace{-20pt}
\caption{Auto-tuning enables faster racing. The baseline (blue) brakes conservatively at each turn. The tuned MPC (yellow) drives much faster.}
\label{fig:racing:tuning}
\vspace{-10pt}
\end{figure}

\textbf{Auto-tuned MPC parameters enable faster racing}: We select the best BO weights and evaluate them on the vehicle, comparing against a baseline using hand-tuned weights.  
Figure~\ref{fig:racing:tuning} shows that the auto-tuned weights yield significantly faster racing. The vehicle's speed $v$ is $2.44~\rm{m/sec.}$ faster in average, and the vehicle turns with higher yaw rates $r$. The total torque $\tau_{\textrm{total}}$ is also larger in average, signaling more aggressive driving.
This automatic tuning experiment exposes a weakness of the baseline MPC, which tracks the reference control trajectory too closely
and brakes at each turn even when the car could drive faster, while \solverName's auto-tuned weights enable driving closer to the limits of handling.

The baseline and \texttt{Optuna}-tuned MPC 
parameters are  in Table~\ref{tab:mpc_params}. 
The higher terminal time cost weight $\alpha$ and lower control reference tracking weights $(R_{\tau_{\textrm{rear}}},R_{\tau_{\textrm{front}}})$ enable more aggressive driving and deviating from the reference control trajectory if it enables faster racing. Indeed, Figure~\ref{fig:racing:tuning} shows harder braking before each turn and earlier accelerations in the turns using tuned weights.  Also, the higher sideslip and angle deviation weights $(Q_\beta,Q_{\Delta\varphi})$ reduce sideslip and enable better acceleration in the turns. 
Finally, the tire friction parameters $(\mu_{\textrm{front}},\mu_{\textrm{rear}})$ change slightly, but this change is likely negligible compared to changes in the cost weights (e.g. compare with the larger changes in \cite{adabag2025differentiable}). 

\subsection{Scalability to longer planning horizons}\label{sec:racing:scaling}
Next, we compare \solverName~against an SQP solver using \OSQP~\cite{Stellato2020} as the inner QP solver. This solver is a strong  baseline, as its SQP outer loop forms the QP approximations on the GPU, and \OSQP uses the multi-threaded Intel MKL PARDISO sparse linear system solver\footnote{OSQP-CUDA is known to be performant only for problems substantially larger than ours~\cite{Schubiger2020, adabag2024mpcgpu}, so we use the CPU implementation of OSQP.}. Both solvers use the same MPC problem formulation, auto-tuned weights from the previous section, and solver hyperparameters, isolating the effect of the inner QP solver on scalability.

\begin{figure}[t]
\centering
\includegraphics[width=1.00\columnwidth]{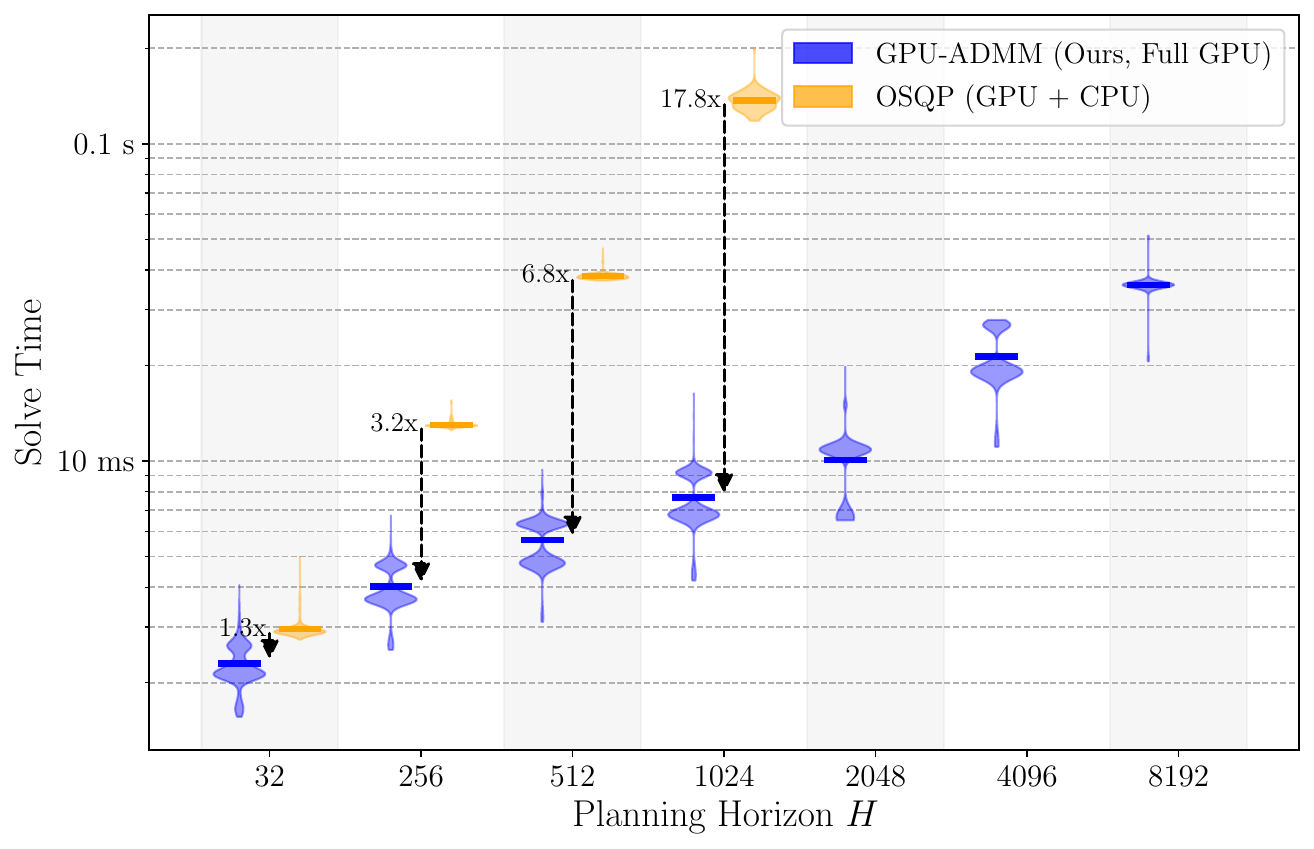}
\vspace{-20pt}
\caption{Solve times across planning horizons $N$, demonstrating the scalability of \solverName~to long horizons beyond what the OSQP baseline supports.}
\vspace{-10pt}
\label{fig:racing:solvetimes}
\end{figure}

\begin{figure}[!t]
\centering
\includegraphics[width=1.00\columnwidth]{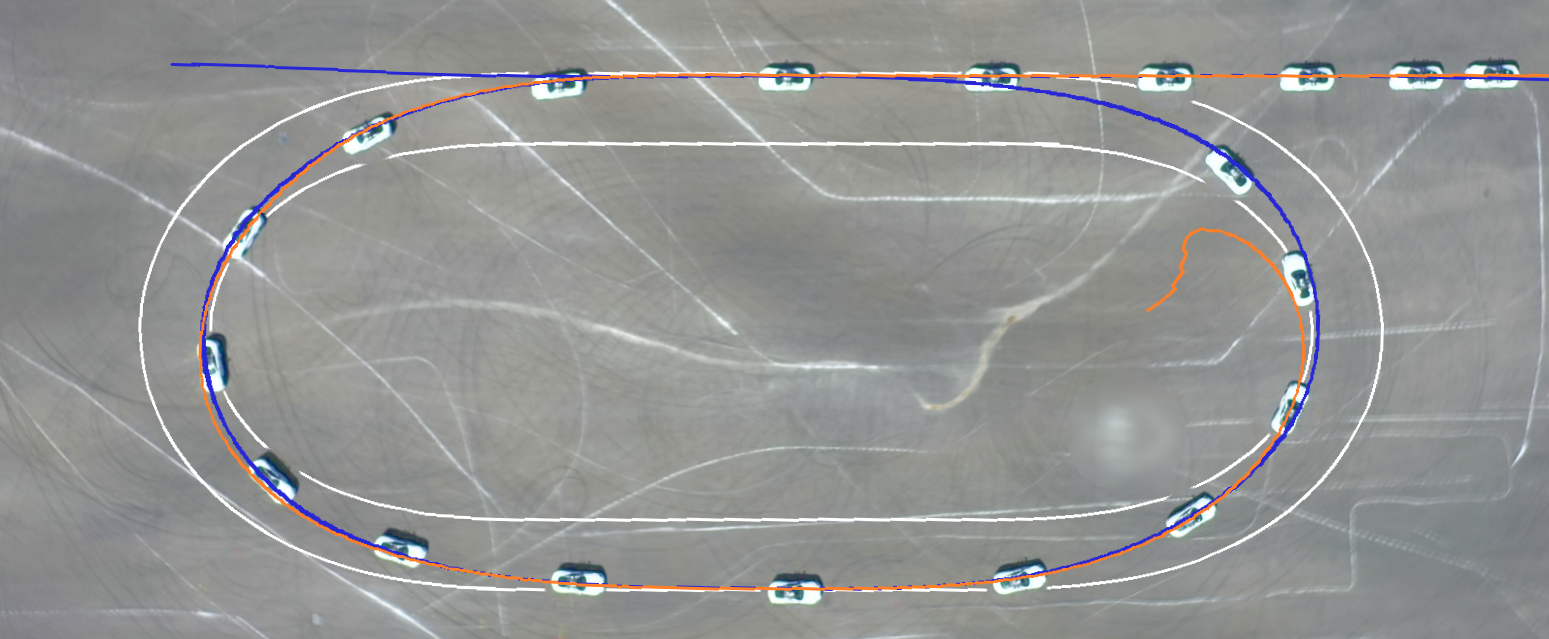}
\includegraphics[width=1.0\columnwidth]{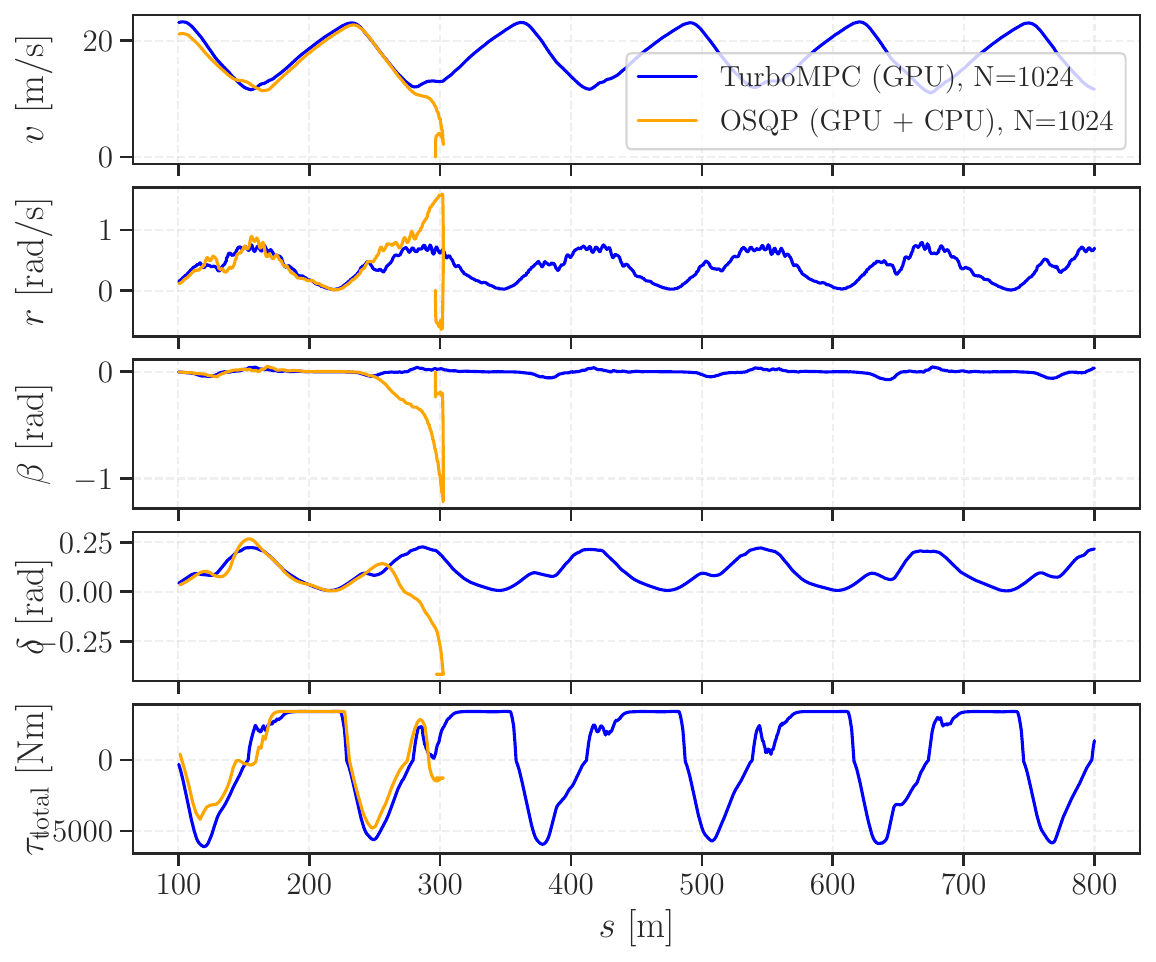}
\vspace{-20pt}
\caption{At a planning horizon  $N=1024$, the \OSQP baseline (yellow) loses control of the vehicle, while \solverName~(blue) drives the vehicle successfully throughout the track.}

\label{fig:racing:solvetimes:spinoutosqp}
\end{figure}

Results in Figure~\ref{fig:racing:solvetimes} show that the OSQP-based baseline is slower than \solverName{}  for longer planning horizons, with the gap widening at longer horizons. Thus, running the solver entirely on the GPU unlocks significant speedups. For example, at $N=1024$ the OSQP-based controller loses control of the vehicle (Fig.~\ref{fig:racing:solvetimes:spinoutosqp}), while \solverName~maintains control. 
On the other hand, \textbf{\solverName~scales to planning horizons of $N>8000$ discretization nodes}, enabling the controller to anticipate multiple corners ahead within a single MPC solve. These results demonstrate  new capabilities in online, very-long-horizon planning.

\section{Discussion, Limitations, and Future Work}\label{sec:conclusion}
Differentiable GPU solvers help integrate MPC into learning pipelines and scale to higher-dimensional systems, longer planning horizons, larger batch sizes,  and more expressive objectives, dynamics, and constraints models such as neural networks. However, \solverName{} comes with limitations.

First, direct sparse factorization methods for solving linear systems pay a constant-factor solve-time overhead relative to iterative methods  such as PCG\cite{adabag2024mpcgpu,adabag2025differentiable}. A hybrid backend that selects PCG for the forward pass and direct factorization elsewhere could recover this advantage.

Second, \solverName{} supports pointwise inequality constraints, linearized as slab constraints \scalebox{0.85}{$\underline{G}\leq Gx\leq\overline{G}$} inside the solver. 
Extending support to conic and complementary constraints would be useful to many problems in robotics (e.g., contact-rich humanoid locomotion without  simplified friction-cone box constraints as in our IL experiments), aerospace (e.g., rocket engine thrust pointing constraints), and others.

Third, \solverName{}'s backward pass inherits the advantages but also the limitations of implicit differentation: (1) Gradient discontinuities at active-set transitions can destabilize training and require small learning rates that slow down learning. 
This is particularly a challenge for long-horizon problems like the racing case where such numerical instabilities can compound over time.
(2) The gradients' accuracy relies on the convergence of the forward pass that solves the MPC problem.  However, gradient-based optimization may not always converge, motivating the development of robust warm-starting strategies. % 
(3) Implicit differentiation requires constraint Hessian information, which can be expensive to obtain. Better understanding where approximate second-order information is sufficient for successful application of differentiable MPC \cite{amos2018differentiable,karkus2023diffstack,adabag2025differentiable} (our code supports this approximation to help further investigation), where it is necessary \cite{frey2025differentiable},   and how second-order information can be efficiently computed, is of interest for future work.

Finally, \solverName{} currently relies on double precision for numerical robustness. This limits how much it can exploit modern GPUs, which increasingly devote more hardware resources to low-precision compute units for deep learning. 
We observed that lower precision yields higher numerical errors and degrades solver convergence and gradient accuracy.
Future work should co-design problem scaling and mixed-precision linear algebra~\cite{arrizabalaga2026differentiable,yilmaz2025roboprec}, so that the solver can better exploit low-precision throughput. % 

\section{Conclusion}\label{sec:conclusion}

\solverName~is a differentiable and GPU-accelerated MPC framework that supports state and control inequalities, implicit integrators, cross-time-coupled costs, and slack variables. The solver combines an SQP outer loop with a custom ADMM inner solver, implicit differentiation, and a co-designed JAX-CUDA implementation. 
We validated \solverName{} in simulation across constrained planning, reinforcement learning, imitation learning, and neural network MPC tasks.
We also deployed \solverName{} on a Lexus LC500 autonomous race car, where GPU-batched Bayesian optimization was used to tune controller parameters to drive closer to the vehicle's limits. The solver enabled planning over horizons with more than $8000$ discretization nodes. 
These results point toward differentiable MPC as a fast, GPU-native, constraint-aware primitive that exploits optimal control structure and can be trained at scale.

\appendix[Mathematical Details of the Solver and Additional Experiment Information]
\label{apdx:solver}
We describe the solver in further details as follows.

\begin{enumerate}
\item[]% 
\hspace{-5mm}\hyperref[apdx:ocp]{\ref{apdx:ocp}\hspace{2mm} Optimal Control Problem}
\hspace*{\fill} 
\pageref{apdx:ocp}
\item[]% 
\hspace{-5mm}\hyperref[apdx:qp]{\ref{apdx:qp}\hspace{2mm} QP Approximation}
\hspace*{\fill} 
\pageref{apdx:qp}
\item[]% 
\hspace{-5mm}\hyperref[apdx:admm]{\ref{apdx:admm}\hspace{2mm} ADMM}
\hspace*{\fill} 
\pageref{apdx:admm}
\hspace{-5mm}\hyperref[apdx:linesearch]{\ref{apdx:linesearch}\hspace{2mm} Line Search}
\hspace*{\fill} 
\pageref{apdx:linesearch}
 
\item[]% 
\hspace{-5mm}\hyperref[apdx:backward]{\ref{apdx:backward}\hspace{2mm} Backward Pass}
\hspace*{\fill} 
\pageref{apdx:backward}
\end{enumerate}  

We then provide additional details on our experiments.
\begin{enumerate}
\item[]% 
\hspace{-5mm}\hyperref[appendix:linear]{\ref{appendix:linear}\hspace{2mm} Linear-System Scaling}
\hspace*{\fill} 
\pageref{appendix:linear}
\item[]% 
\hspace{-5mm}\hyperref[appendix:humanoid]{\ref{appendix:humanoid}\hspace{2mm} Humanoid Balancing Imitation Learning}
\hspace*{\fill} 
\pageref{appendix:humanoid}
\item[]% 
\hspace{-5mm}\hyperref[app:pointmass-rl]{\ref{app:pointmass-rl}\hspace{2mm} 3D Point-Mass Tracking Training}
\hspace*{\fill} 
\pageref{app:pointmass-rl}
\end{enumerate}

\subsection{Optimal Control Problem (\OCP)}
\label{apdx:ocp}
We define the \OCP in Sec.~\ref{sec:Contribution} and restate it below:
\begin{subequations}
    \begin{align} \nonumber
&\hspace{3cm}\OCP
\\[-2mm]
\min_{x,u,\xi}\ &
\sum_{t=0}^{N-1}\ell_t(x_t,...,u_{t+1})
+\ell_N(x_N,u_N) 
 +\tfrac{\gamma_\xi}{2}
\sum_{t=0}^{N}\|\xi_t\|_2^2,\\
\text{s.t.}\quad &x_0=x_{\rm init},\\  
&f_t(x_t,u_t,x_{t+1},u_{t+1})=0,\quad t=0,\ldots,N-1,\\  
&\underline g_t\le g_t(x_t,u_t)+\delta_\xi\xi_t\le \overline g_t,  
  \quad t=0,\ldots,N.
\end{align}
\end{subequations}
where 
$x_t\in\R^{n_x}$ and $u_t\in\R^{n_u}$ are the state and control at node $t=0,\ldots,N$, and the binary $\delta_\xi\in\{0,1\}$ toggles the use of slack variables $\xi_t$, penalized by the scalar penalty $\gamma_\xi>0$.

All the \OCP functions are assumed twice continuously differentiable in $(x,u)$.

\begin{tcolorbox}[
title={% 
\textbf{Remark A\customlabel{remark:dynamics_implicit}{1}}(Dynamics constraints \& implicit integrators)
},
coltitle=black, 
boxrule=2pt,
colframe=black!6, % 
enhanced, colback=black!2, boxsep=0pt, left=6pt, right=6pt]
The dynamics equality constraints support both explicit and implicit integrators.  Explicit dynamics constraints do not depend on $u_{t+1}$ and can be written as
$$
    f_t(x_t,u_t,x_{t+1},u_{t+1})
    =\Phi_t(x_t,u_t)-x_{t+1}.
$$
For a trapezoidal
implicit scheme with continuous-time dynamics $\dot{x}_t=f^\mathrm{ct}_{t}$ and step size
$\Delta t$,
$$
f_t(\dots) 
=x_t+\tfrac{\Delta t}{2}(f^\mathrm{ct}_{t}(x_t,u_t)+f^\mathrm{ct}_{t+1}(x_{t+1},u_{t+1}))-x_{t+1}.
$$
\end{tcolorbox}

\subsection{QP Approximation}\label{apdx:qp}
At each SQP iterate $(\bar x,\bar u)$, we form a quadratic approximation of the
objective and linear approximations of the constraints, yielding the 
quadratic program (QP) defined in Sec.~\ref{sec:solver:qp} and restated below:
\begin{subequations}\label{eq:qp}
\begin{align}
\hspace{-1cm}\QP:
\quad
\min_{(x,\xi)}\quad&
\tfrac12 x\trans P x + q\trans x + \tfrac{\gamma_\xi}{2}\|\xi\|^2
\\
\text{s.t.}\quad\,&
C x = c,
\ \ \ 
\underline{G} \le Gx +\delta_\xi\xi
   \le \overline{G}.
\end{align}
\end{subequations}
where $x$  
stacks the  stage variables
$
x:=(x_t,u_t)_{t=0}^N
$. 

The $P$, $C$, and $G$ matrices derived from the cost and constraint approximations, are block-tridiagonal, -bidiagonal, and -diagonal, respectively, with dimensions denoted below. 
\begin{itemize}[leftmargin=5mm]
\item $P\succeq 0$ contains the Hessian of the cost quadratic approximation.
\begin{equation}
P=
\AverageSmallMatrix{
D_0 & E_0^\top \\[-2mm]
E_0 & D_1 & \ddots \\[-2mm]
& \ddots & \ddots & E_{N-1}^\top\\
& & E_{N-1} & D_N
}
\in\mathbb{R}^{(N+1)n\times (N+1)n}.
\end{equation}
\item $C$ encodes the linearized initial and dynamics constraints via $C x = c$.
\begin{equation}
C=
\AverageSmallMatrix{
A_{\mathrm{init}} & 0 & 0 & \cdots & 0 \\
A_0^{-} & A_0^{+} & 0 & \cdots & 0 \\[-1mm]
0 & A_1^{-} & A_1^{+} & \ddots & \vdots \\[-1.5mm]
\vdots & \ddots & \ddots & \ddots & 0 \\
0 & \cdots & 0 & A_{N-1}^{-} & A_{N-1}^{+}
}\in\R^{(n_{\mathrm{init}}+Nn)\times (N+1)n}.
\end{equation}
\item $G$ encodes the linearized inequality constraints, with optional slack variables, via $\underline{G} \le Gx +\delta_\xi\xi
   \le \overline{G}$:
\begin{equation}
G =
\AverageSmallMatrix{
G_0 & 0 & 0 \\[-2mm]
0 &  \ddots & 0 \\
0   &  0 &   G_N
}\in\R^{(N+1)n_g\times (N+1)n}.
\end{equation}
\end{itemize}
Leveraging the sparse structure of $P$, $C$, and $G$ is key to an efficient implementation. The following sections describe other terms in further detail.

\subsubsection{Cost} 
For each stage \(t=0,\dots,N-1\), we quadratize \(\ell_t\) around \((\bar x_t,\bar x_{t+1})\) ($x_t\gets(x_t,u_t)$ denotes the stacked variable):  
\begin{align}
\ell_t(x_t,x_{t+1})
\;\approx\;&
\ell_t(\bar x_t,\bar x_{t+1})
+ q_t^{-\top}\delta x_t
+ q_t^{+\top}\delta x_{t+1}
\nonumber\\
& 
\hspace{-3mm}
+ \tfrac12
\begin{bmatrix}\delta x_t\\ \delta x_{t+1}\end{bmatrix}^\top
\begin{bmatrix}
H_t^{-} & H_t^{\pm}\\
H_t^{\pm\top} & H_t^{+}
\end{bmatrix}
\begin{bmatrix}\delta x_t\\ \delta x_{t+1}\end{bmatrix},
\end{align}
where $\delta x:=x-\bar x$ and 
\begin{gather*}
q_t^{-} := \nabla_{x_t}\ell_t,\qquad
q_t^{+} := \nabla_{x_{t+1}}\ell_t,
\\
H_t^{-} := \nabla^2_{x_t x_t}\ell_t,\quad
H_t^{\pm} := \nabla^2_{x_t x_{t+1}}\ell_t,\quad
H_t^{+} := \nabla^2_{x_{t+1} x_{t+1}}\ell_t,
\end{gather*}
all evaluated at \((\bar x_t,\bar x_{t+1})\).
Similarly for the terminal cost,
\begin{equation}
\ell_N(x_N)\approx
\ell_N(\bar x_N)
+ q_N^{-\top}\delta x_N
+ \tfrac12 \delta x_N^\top H_N^{-}\,\delta x_N,    
\end{equation}
with \(q_N^{-\top}:=\nabla_{x_N}\ell_N(\bar x_N)\) and \(H_N^{-}:=\nabla^2_{x_N}\ell_N(\bar x_N)\).

Collecting all terms, the block entries of $P$ are
\begin{subequations}
    \begin{align}
D_0 &= H_0^{-},\\
D_t &= H_{t-1}^{+}+H_t^{-}, \qquad &&t=1,\dots,N,
\\
E_t &= H_t^{\pm},\qquad &&t=0,\dots,N-1,
\end{align}
\end{subequations}
and the linear term $q=(q_0,\dots,q_N)\in\R^{(N+1)n}$, with
\begin{align}
q_0 &= q_0^{-}, \qquad q_t = q_{t-1}^{+} + q_t^{-}, \quad t=1,\dots,N.
\end{align}

\subsubsection{Equality constraints}
The initial state and (optional) control constraints are already linear and define $A_{\mathrm{init}}$ in $C$. The dynamics constraints are linearized and take the form 
\begin{equation}
    A^-_t
    \begin{bmatrix}x_t\\ u_t\end{bmatrix}
    +A^+_t
    \begin{bmatrix}x_{t+1}\\ u_{t+1}\end{bmatrix}
    =b_t,\quad t=0,\ldots,N-1,
\end{equation}
where $A^-_t$ and $A^+_t$ are Jacobian blocks of
$f_t(x_t,u_t,x_{t+1},u_{t+1})$ at
$(\bar x_t,\bar u_t,\bar x_{t+1},\bar u_{t+1})$, and 
\begin{equation}
b_t
=
A^-_t
\begin{bmatrix}\bar x_t\\ \bar u_t\end{bmatrix}
+A^+_t
\begin{bmatrix}\bar x_{t+1}\\ \bar u_{t+1}\end{bmatrix} 
-f_t(\bar x_t,\bar u_t,\bar x_{t+1},\bar u_{t+1}).
\end{equation}
The affine term of the equality constraints is then defined as $c=
(
b_{\mathrm{init}},
b_0,
b_1,
\dots,
b_{N-1}
)$.

\subsubsection{Inequality constraints}
They are linearized as
\begin{equation}
    \underline G_t\le
    G_t\begin{bmatrix}x_t\\ u_t\end{bmatrix}
    +\delta_\xi\xi_t
    \le \overline G_t,\quad
    G_t=\nabla_{(x,u)}g_t(\bar x_t,\bar u_t),
\end{equation}
with shifted bounds
$\underline G_t=\underline g_t-g_t(\bar x_t,\bar u_t)
+G_t\begin{bmatrix}\bar x_t\\ \bar u_t\end{bmatrix}$ and $\overline G_t=\overline g_t-g_t(\bar x_t,\bar u_t)
+G_t\begin{bmatrix}\bar x_t\\ \bar u_t\end{bmatrix}$.
Stacking these constraints for all stages \(t=0,\dots,N\) gives the  block diagonal  matrix $G$ and  the stacked bounds 
$\underline G := (\underline G_0,\dots, \underline G_N)$ and $\overline G := (\overline G_0,\dots, \overline G_N)$.

\subsection{ADMM}
\label{apdx:admm}

We reformulate the \QP as in \OSQP. The main formulation is described in Sec.~\ref{subsec:sqp:admm}. We further contrast the derivations with and without slack variables below.

\subsubsection{ADMM without slack variables (\OSQP, $\delta_\xi=0$)}
\label{apdx:admm:osqp}
First, we derive our ADMM scheme for the case without slack variables $(\delta_\xi,\xi)=(0,0)$. The scheme follows \OSQP.

The corresponding augmented Lagrangian is given by
\begin{align}\label{eq:admm:derivations:lagrangian}
\mathcal L_{\sigma,\rho}(\tilde x,\tilde z,x,z,w,y)
&=
\tfrac12 \tilde x^\top P \tilde x + q^\top \tilde x
+ \mathcal{I}_{[l,u]}(z)
+  \mathcal{I}_0(A\tilde x-\tilde z)
\nonumber\\
&\hspace{-18mm}
+ \tfrac{\sigma}{2}\|\tilde x-x+\sigma^{-1}w\|^2
+ \tfrac\rho2\|\tilde z-z+\rho^{-1}y\|^2.
\end{align}
Then, applying ADMM consists of
\begin{enumerate}[label=\alph*)]
    \item minimizing $\mathcal L_{\sigma,\rho}$ over $(\tilde x, \tilde z)$
    \item minimizing $\mathcal L_{\sigma,\rho}$ over  $(x,z)$
    \item updating the multipliers $(w,y)$
\end{enumerate}
By also including an over-relaxation strategy, these steps produce the following updates.

\textit{\textbf{Primal update:}}
\begin{align}
\nonumber
(\tilde x^{k+1}, \tilde z^{k+1})
\leftarrow
&\argmin_{(\tilde x,\tilde z)} 
\label{eq:admm:primal}
 \tfrac12 \tilde x^\top P \tilde x {+} q^\top \tilde x
{+} \tfrac{\sigma}{2}\|\tilde x{-}x^k{+}\sigma^{-1}w^k\|^2
\nonumber
\\
&\hspace{-3mm} 
+ \tfrac\rho2\|\tilde z-z^k+\rho^{-1}y^k\|^2
\ \ \text{s.t.}\ \ A\tilde x=\tilde z.
\end{align}

\textit{Over-relaxation:} Given $\alpha\in(0,2)$, let
$$
\hat{x}^{k+1}=\alpha \tilde x^{k+1}+(1-\alpha)x^k, \ 
\hat{z}^{k+1}=\alpha \tilde z^{k+1}+(1-\alpha)z^k.
$$

\textit{\textbf{Slack update:}}
\begin{subequations}\label{eq:admm:derivations:slack_updates_full}
\begin{align}
x^{k+1}
&\leftarrow
\operatorname*{arg\,min}_{x}\ 
\tfrac{\sigma}{2}\|\hat x^{k+1}-x+\sigma^{-1}w^k\|^2
\nonumber
\\
&= \hat x^{k+1} + \sigma^{-1}w^k,
\label{eq:admm:derivations:slack_x}
\\
z^{k+1}
&\leftarrow
\operatorname*{arg\,min}_{z}\ 
\mathcal{I}_{[l,u]}(z) + \tfrac\rho2\|\hat z^{k+1}-z+\rho^{-1}y^k\|^2
\nonumber
\\
&= \Pi_{[l,u]}\!\left(\hat z^{k+1}+\rho^{-1}y^k\right),
\label{eq:admm:derivations:slack_z}
\end{align}
\end{subequations}

with $z$ decomposed in:
\begin{subequations}
    \begin{align}
z_f^{k+1} &= c,
\\
z_g^{k+1} 
&= 
\Pi_{[\underline{G},\overline{G}]}\!\left(\alpha G\tilde x^{k+1}+(1-\alpha)z_g^k+\rho_g^{-1}y_g^k\right).
\end{align}
\end{subequations}

Because $z_f^k\equiv c$ for all ADMM steps, our implementation does not store $z_f$ and substitutes it with $c$ throughout.

\textit{\textbf{Dual update:}}
\begin{subequations}\label{eq:admm:derivations:dual_updates_full}
\begin{align}
w^{k+1} &\leftarrow w^k + \sigma(\hat x^{k+1}-x^{k+1}),
\label{eq:admm:derivations:dual_w}
\\
y^{k+1} &\leftarrow y^k + \rho(\hat z^{k+1}-z^{k+1}),
\label{eq:admm:derivations:dual_nu}
\end{align}
\end{subequations}
\qquad with $y$ decomposed in:
\begin{align}
y_f^{k+1} &= y_f^k + \rho_f\bigl(\alpha C\tilde x^{k+1}+(1-\alpha)z_f^k-c\bigr), 
\\
y_g^{k+1} &= y_g^k + \rho_g\bigl(\alpha G\tilde x^{k+1}+(1-\alpha)z_g^k-z_g^{k+1}\bigr). 
\end{align}

\subsubsection*{Eliminating $w$ and $\hat x$}
From \eqref{eq:admm:derivations:slack_x} and \eqref{eq:admm:derivations:dual_w},
\begin{equation}
w^{k+1}
=
w^k + \sigma\bigl(\hat x^{k+1}-(\hat x^{k+1}+\sigma^{-1}w^k)\bigr)=0,    
\end{equation}
so $w^k= 0$ and
$x^k = \hat x^k$ for all $k\ge 1$.
Substituting into the ADMM updates above and unrolling the over-relaxation step yields Algorithm \ref{alg:admm}.

\textit{\textbf{Linear-system solve via the Schur complement method:}}
The primal update \eqref{eq:admm:primal} is an equality-constrained QP. 
Its KKT conditions  are
\begin{subequations}\label{eq:admm:kkt_primal_subproblem}
\begin{align}
(P+\sigma I)\tilde x^{k+1} + q - \sigma x^k + A^\top \nu^{k+1} &= 0,
\label{eq:admm:kkt_primal_x}
\\
\rho \tilde z^{k+1} - \rho z^k + y^k - \nu^{k+1} &= 0,
\label{eq:admm:kkt_primal_z}
\\
A\tilde x^{k+1} - \tilde z^{k+1} &= 0,
\label{eq:admm:kkt_primal_eq}
\end{align}
\end{subequations}
where $\nu^{k+1}$ denotes the multipliers associated with the equality constraint \(A\tilde x=\tilde z\).

Eliminating 
$
\tilde z^{k+1}\mathop{=}\limits^{\eqref{eq:admm:kkt_primal_z}}
z^k+\rho^{-1}(\nu^{k+1}-y^k)$, 
and stacking \eqref{eq:admm:kkt_primal_x} and \eqref{eq:admm:kkt_primal_eq} gives the linear system
\begin{equation}\label{eq:KKT}
\begin{bmatrix}
P+\sigma I & A^\top\\
A & -\rho^{-1}I
\end{bmatrix}
\begin{bmatrix}
\tilde x^{k+1}\\ \nu^{k+1}
\end{bmatrix}
=
\begin{bmatrix}
\sigma x^k-q\\
z^k-\rho^{-1}y^k
\end{bmatrix}.
\end{equation}
From the second block row,
we obtain
\begin{equation}\label{eq:admm:y_update}
\nu^{k+1}=y^k+\rho\bigl(A\tilde x^{k+1}-z^k\bigr),
\end{equation}
which once substituted into \eqref{eq:admm:kkt_primal_x} yields the Schur complement system $S\tilde x^{k+1}=\eta^k$ defined in~\eqref{eq:admm:schur}.

After solving the linear system, we recover
\begin{equation*}
\tilde z^{k+1}\mathop{=}^\eqref{eq:admm:kkt_primal_eq}A\tilde x^{k+1}.
\end{equation*}

\textit{\textbf{Structure from OCP:}}
We specialize to the OCP QP structure by partitioning
\[
A=\begin{bmatrix}C\\G\end{bmatrix}\hspace{-1mm},\ 
l=\begin{bmatrix}c\\\underline{G}\end{bmatrix}\hspace{-1mm},
u=\begin{bmatrix}c\\\overline{G}\end{bmatrix}\hspace{-1mm},
z=\begin{bmatrix}z_f\\z_g\end{bmatrix}\hspace{-1mm}, 
\tilde z=\begin{bmatrix}\tilde z_f\\\tilde z_g\end{bmatrix}\hspace{-1mm}, 
y=\begin{bmatrix}y_f\\y_g\end{bmatrix}\hspace{-1mm},
\]
and choosing 
\begin{equation*}\label{eq:rho_def}
\rho=\diag(\rho_f I,\rho_g I)
\end{equation*}
to reflect the  equality and inequality constraints.

Hence, the primal update \eqref{eq:admm:primal} consists of the steps
\begin{enumerate}
\item $\tilde x^{k+1}\gets \mathop{\text{Solve}}\limits_x\{Sx^{k+1}=\eta^k\}$,
\item $(\tilde z_f^{k+1},\tilde z_g^{k+1})
=
(C\tilde x^{k+1},G\tilde x^{k+1})$,
\end{enumerate}
using the Schur complement components 
\begin{align}
S &= P+\sigma I + C^\top\rho_f C + G^\top\rho_g G
\\ \nonumber
&=
\begin{bmatrix}
\Theta_0 & \Phi_0^\top \\
\Phi_0 & \Theta_1 & \ddots \\
& \ddots & \ddots & \Phi_{N-1}^\top \\
& & \Phi_{N-1} & \Theta_N
\end{bmatrix},
\\
\eta^k &= \sigma x^k-q
+ C^\top(\rho_f z_f^k-y_f^k)
+ G^\top(\rho_g z_g^k-y_g^k),
\end{align}
with
\begin{small}
\begin{align*}
\Theta_0
&=
D_0 + \sigma I
+ \rho_f(A_{\rm init}^\top A_{\rm init}
+ A_0^{-\top} A_0^-)
+ \rho_g G_0^\top G_0,
\\
\Theta_t
&=
D_t + \sigma I
+ \rho_f(A_{t-1}^{+\top}A_{t-1}^+
+ A_t^{-\top}A_t^-)
+ \rho_g G_t^\top G_t,
\\
&\hspace{4cm}
t=1,\dots,N-1,
\\
\Theta_N
&=
D_N + \sigma I
+ \rho_f A_{N-1}^{+\top}A_{N-1}^+
+ \rho_g G_N^\top G_N.
\\
\Phi_t
&=
E_t + \rho_f A_t^{+\top}A_t^-,
\qquad\ \ \,  t=0,\dots,N-1,
\end{align*}
\end{small} 

\subsubsection{ADMM with slack variables (\texttt{ADMMSlack}, $\delta_\xi=1$)}\label{apdx:admm:admmslack}

We tackle problems with slack variables as follows. 
While the \QP with $\delta_\xi=1$ is a quadratic program that could be handled via the \OSQP scheme presented previously, this approach is more complicated to implement: the structure of the primal linear system would change and one would need to keep track of the slack variables $\xi$.

Instead, we adapt ideas from the \texttt{ADMMSlack}~\cite{LewSlackADMM2025} approach to tackle the \QP previously defined by~\eqref{eq:admmslack}: 
\begin{align*}
\min_{(x,z,\tilde x,\tilde z,\xi)}\, &
\tfrac12 \tilde x^\top P \tilde x \,{+}\, q^\top \tilde x
\,{+}\, \tfrac{\gamma_{\xi}}{2}\|\xi\|^2
\,{+}\, I_{\{c\}}(z_f)
\,{+}\, I_{[\underline G,\overline G]}(z_g\,{+}\,\xi)
\nonumber
\\
\text{s.t.}\quad&
A\tilde x-\tilde z=0,\quad \tilde x-x=0,\quad \tilde z-z=0,
\end{align*}
with augmented Lagrangian defined before by~\eqref{eq:admm:aug_Lagrangian}: 
\begin{align*}
\mathcal L_{\sigma,\rho}(\tilde x,\tilde z,x,z,\xi,w,y)
&=
\tfrac12 \tilde x^\top P \tilde x + q^\top \tilde x
+ \tfrac{\gamma_{\xi}}{2}\|\xi\|^2+
\nonumber
\\
&\hspace{-24mm}
+ I_{\{c\}}(z_f)
+ I_{[\underline G,\overline G]}(z_g+\xi) 
+ I_0(A\tilde x-\tilde z)
\nonumber\\
&\hspace{-24mm}
+ \tfrac{\sigma}{2}\|\tilde x-x+\sigma^{-1}w\|^2
+ \tfrac\rho2\|\tilde z-z+\rho^{-1}y\|^2.
\end{align*}

Then, our ADMM scheme  consists of
\begin{enumerate}[label=\alph*)]
\item minimizing $\mathcal L_{\sigma,\rho}$ over $(\tilde x, \tilde z)$
\item minimizing $\mathcal L_{\sigma,\rho}$ over  $(x,z,\xi)$
\item updating the multipliers $(w,y)$
\end{enumerate} 
These steps give the following updates, including an over-relaxation strategy similar to the case without slacks.

\textit{\textbf{Primal update:}}
Same as in \eqref{eq:admm:primal}

\textit{\textbf{Slack update:}}
Same as in \eqref{eq:admm:derivations:slack_updates_full}, but with
\begin{subequations}
\label{eq:admm:slack:derivations}
\begin{align}
(z_g^{k+1},\xi^{k+1})
&\leftarrow
\operatorname*{arg\,min}_{(z_g,\xi)}
\ 
\tfrac{\gamma_{\xi}}{2}\|\xi\|^2+I_{[\underline G,\overline G]}(z_g+\xi) + 
\nonumber
\\[-2mm]
&\hspace{14mm}+\tfrac\rho2\|\hat z_g^{k+1}-z_g+\rho^{-1}y_g^k\|^2
\label{eq:admm:slack:derivations:1}
\\
\label{eq:admm:slack:derivations:2}
  &=
  \begin{bmatrix}
  \widetilde{\Pi}^{\gamma_{\xi}}_{[\underline G,\overline G]}\bigl(\hat z_g^{k+1}+\rho_g^{-1}y_g^k\bigr)
  \\
  \Delta\widetilde{\Pi}^{\gamma_{\xi}}_{[\underline G,\overline G]}\bigl(\hat z_g^{k+1}+\rho_g^{-1}y_g^k\bigr) 
  \end{bmatrix},
\end{align}
\end{subequations}
\qquad where $\widetilde{\Pi}$ and $\Delta\widetilde{\Pi}$ are  defined in \eqref{eq:proj:tilde}.

\textit{\textbf{Dual update:}}
Same as \eqref{eq:admm:derivations:dual_updates_full}

The step in \eqref{eq:admm:slack:derivations} follows from \cite[Lemma 1]{LewSlackADMM2025}. Specifically, denoting $Z=\hat z_g^{k+1}+\rho^{-1}_gy_g^k$, with appropriate substitutions, the last result of the proof of \cite[Lemma 1]{LewSlackADMM2025} shows 
\begin{align*}
&\begin{cases}
z_i=\frac{\rho Z_{i}+\gamma_{\xi}\underline G_i}{\rho+\gamma_{\xi}},
\ \ 
\xi_i =\tfrac{\rho}{\rho+\gamma_{\xi}}(\underline G_i-Z_i)
&\text{if }Z_i<\underline G_i,
\\ 
z_i=\frac{\rho Z_i+\gamma_{\xi} \overline G_i}{\rho+\gamma_{\xi}},
\ \ 
\xi_i =\tfrac{\rho}{\rho+\gamma_{\xi}}(\overline G_i-Z_i),
&\text{if }Z_i> \overline G_i,
\\ 
z_i=Z_i,
\ \ 
\xi_i=0
&\text{if }\underline G_i\leq Z_i\leq \overline G_i,
\end{cases}
\end{align*}
and \eqref{eq:admm:slack:derivations:2} then follows from the definitions of $(\widetilde{\Pi},\Delta\widetilde{\Pi})$.

The remainder of the derivations to retrieve the steps in Algorithm \ref{alg:admm} follow those for the case without slack variables as described previously.  
\\
\begin{tcolorbox}[
title={% 
\textbf{Remark A\customlabel{remark:ADMMSlackvsOSPQ}{2}}(\texttt{ADMMSlack}: Differences with \OSQP)
},
coltitle=black, 
boxrule=2pt,
colframe=black!6, % 
enhanced, colback=black!2, boxsep=0pt, left=6pt, right=6pt]
The \OSQP  ADMM scheme  would introduce a copy $\tilde\xi$ of $\xi$, and use a primal  update that minimizes the Lagrangian over $(\tilde x, \tilde z, \tilde\xi)$, solving a larger linear system. Instead, our ADMM scheme   for handling slack variables is inspired from \cite{LewSlackADMM2025}. It is easier to implement as it only consists of a small change to the projection step. 
\end{tcolorbox}

\subsubsection{Step-size parameters selection}\label{apdx:admm:rho}
The performance of ADMM highly depends on the choice of the step-size parameters $(\sigma,\rho)$. 
As in \cite{Stellato2020}, we fix $\sigma=10^{-6}$.   
We also partition $\rho=\operatorname{diag}(\rho_f I,\rho_g I)$ to reflect the always-active equality block and the inequality block and  set
$$
\rho_f = 10^3\bar\rho,
\qquad
\rho_g=\bar\rho,
$$
with initial $\bar\rho=0.1$. We adapt $\bar\rho$ to balance primal and dual residuals following an adaptive schedule as described in Sec.~\ref{subsec:sqp:admm}.

\subsection{Line search}\label{apdx:linesearch}
We use a standard backtracking linesearch~\cite[Chapter~18]{Nocedal2006}. After each \QP subproblem solve formulated at $\tilde{x}=(x,\xi)$ and with solution 
$\tilde{x}^*=(x^*,\xi^*)$, we search along the direction $\Delta \tilde{x}=\tilde{x}^*-\tilde{x}$.  We use the merit function
\begin{equation*}
\scalemath{0.9}{\varphi(\tilde{x}) = \ell(\tilde{x}) + \mu\left(\|f(\tilde{x})\|_1 + \|[g(\tilde{x})-\bar g]_+\|_1
+ \|[\underline g-g(\tilde{x})]_+\|_1\right)},
\end{equation*}
where $\ell(\tilde{x})$ is the total cost, $f(\tilde{x})$ corresponds to equality constraints, and
$g(\tilde{x})$ to inequality constraints.  The penalty parameter is
chosen as
\begin{equation*}
\scalemath{0.9}{\mu =
\frac{\nabla \ell(\tilde{x})^\top \Delta \tilde{x}}
{\tfrac{1}{2}\max\left(\|h(\tilde{x})\|_1 + \|[g(\tilde{x})-\bar g]_+\|_1
+ \|[\underline g-g(\tilde{x})]_+\|_1,\epsilon_c\right)}}.
\end{equation*}
A candidate step size
$\alpha$ is accepted if it satisfies the Armijo-type condition
\begin{equation*}
\varphi(\tilde{x}+\alpha\Delta \tilde{x})
\leq
\varphi(\tilde{x}) + \eta\alpha D_\mu, 
\end{equation*}
with the directional derivative $D_\mu =
\nabla \ell(\tilde{x})^\top \Delta \tilde{x}
-
\mu\left(\|f(\tilde{x})\|_1 + \|[g(\tilde{x})-\bar g]_+\|_1
+ \|[\underline g-g(\tilde{x})]_+\|_1\right)$ and  $\eta=0.4$.  Merit values are evaluated in parallel over a set of $\alpha$ values and the largest candidate satisfying the
decrease condition is selected.  If no finite candidate satisfies the Armijo condition,
we take the finite candidate with the smallest merit value.

\subsection{Backward Pass: Sensitivities Computation}
\label{apdx:backward}

The solution $(x_\theta,\xi_\theta)$ to \OCP depends on parameters
$\theta\in\R^p$ through the cost and constraints.  To integrate the solver into
learning pipelines, we compute gradients of functions of the solution
\begin{equation}
    \frac{\partial \mathcal L(x_\theta,\xi_\theta)}{\partial \theta},
\end{equation}
where $\mathcal L$ denotes a downstream scalar loss.  We use implicit differentiation, computing these
gradients through differentiation of the KKT conditions at the converged primal-dual
solution, while keeping the active set fixed.

In the following, all quantities are evaluated at the
converged primal-dual solution, and we suppress explicit dependence on
$\theta$ whenever this does not create ambiguity.

\subsubsection{Active constraints identification}
Let $f(x)=0$ collect all initial and dynamics equality constraints. Let $\mathcal A=\underline {\mathcal A} \cup \overline{\mathcal A}$ denote the set of active inequality
constraints. An inequality constraint belongs to $\underline {\mathcal A}$ (i.e., it is lower active) if 
\begin{equation*}
    \scalemath{0.95}{y_{g_i}<-\epsilon_g~\text{or}~(|y_{g_i}|<\epsilon_g~\text{and}~g_i(x)-\underline g_i \leq \epsilon_g|\underline g_i|)},
\end{equation*}
and it belongs to $\overline {\mathcal A}$ (i.e., it is upper active) if
\begin{equation*}
    \scalemath{0.95}{y_{g_i}>\epsilon_g~\text{or}~(|y_{g_i}|<\epsilon_g~\text{and}~\overline g_i-g_i(x) \leq \epsilon_g|\overline g_i|)},    
\end{equation*}
where $\epsilon_g>0$ is a user-defined tolerance, with our code using a combination of $\epsilon_{\textrm{abs}}$ and $\epsilon_{\textrm{rel}}$. The dual check gives the KKT-consistent active side and takes priority, while the proximity to the bounds is used only when the dual side is ambiguous (i.e., $|y_{g_i}|<\epsilon_g$). From this active-set check, we identify the active constraints as $g_{\mathcal A}$.

In the presence of  slack variables ($\delta_\xi=1$), the active constraints are written compactly as
\begin{equation}
    g_{\mathcal A}(x)+S\xi_{\mathcal A}=0,
    \qquad
    S=\operatorname{diag}(S_i),
\end{equation}
where
\[
S_i=
\begin{cases}
+1, & \text{if the lower bound is active or $\xi_i=0$},\\
-1, & \text{if the upper bound is active},
\end{cases}
\]
so that $S_i=\operatorname{sign}(\xi_i)$ for active softened constraints.
Inactive softened constraints have $\xi_i=0$ and are not included in the
backward KKT system.

\subsubsection{Implicit differentiation} We compute the gradients by differentiating the KKT conditions. 
We denote the active-set
Lagrangian by
\begin{equation}
    L(x,\xi_{\mathcal A},y)
    =
    \ell(x)
    +
    \tfrac{\gamma_{\xi}}{2}\|\xi_{\mathcal A}\|^2
    +
    y_f^\top f(x)
    +
    y_g^\top\bigl(g_{\mathcal A}(x)+S\xi_{\mathcal A}\bigr),
\end{equation}
where $y_f$ are the multipliers of the equality constraints and $y_g$ are
the multipliers of the active inequality constraints. 

The \OCP KKT
conditions are
\begin{subequations}
    \begin{align}
    \nabla_x \ell(x)
    +\nabla_x f(x)^\top y_f
    +\nabla_x g_{\mathcal A}(x)^\top y_g
    &=0,\\
    \gamma_{\xi} \xi_{\mathcal A}+Sy_g
    &=0,\\
    f(x)&=0,\\
    g_{\mathcal A}(x)+S\xi_{\mathcal A}&=0.
    \end{align}    
\end{subequations}
We write these conditions compactly as
\begin{equation}
    F(w,\theta)=0,
\qquad
w=(x,\xi_{\mathcal A},y).
\end{equation}

We can now use standard implicit differentiation techniques to obtain gradients. 
By the chain rule, differentiating $F(w(\theta),\theta)=0$ with respect to $\theta$ gives
\begin{equation}
\frac{\partial F}{\partial w}
\frac{\partial w}{\partial \theta}
+
\frac{\partial F}{\partial \theta}
=0
\implies
\label{eq:bwd:ift}
\frac{\partial w}{\partial \theta}
=
-
\left[
\frac{\partial F}{\partial w}
\right]^{-1}
\frac{\partial F}{\partial \theta}.
\end{equation}
 The above computation corresponds to a Jacobian-Vector Product (JVP), and requires solving a large linear system. Instead, the gradient of a loss $\nabla_\theta \mathcal L(x_\theta,\xi_\theta)$ can be more efficiently computed via a Vector-Jacobian Product (VJP):
\begin{align}\label{eq:loss_gradient}
\hspace{-2mm}
\nabla_\theta \mathcal L^\top 
&= 
-\frac{\partial F}{\partial \theta}^\top
\left(
\left[\frac{\partial F}{\partial w} \right]^{-1}
\AverageSmallMatrix{
\nabla_x\mathcal L\\
\nabla_{\xi_{\mathcal A}}\mathcal L\\
0
}
\right)
= 
-\frac{\partial F}{\partial \theta}^\top
\AverageSmallMatrix{
\eta_x\\
\eta_\xi\\ 
\eta_y
},
\end{align}
which also follows from the chain rule, and requires solving the linear system
\begin{equation}
\label{eq:bwd:slack_adjoint}
\frac{\partial F}{\partial w}
\begin{bmatrix}
\eta_x\\
\eta_\xi\\ 
\eta_y
\end{bmatrix}
=
\begin{bmatrix}
\nabla_x\mathcal L\\
\nabla_{\xi_{\mathcal A}}\mathcal L\\
0 
\end{bmatrix},
\end{equation}
with $\eta_y=(\eta_{y_f},\eta_{y_g})$ and 
\begin{equation*}
\frac{\partial F}{\partial w}
=
\AverageSmallMatrix{
H & 0 & \nabla_x f^\top & \nabla_x g_{\mathcal A}^\top\\
0 & \gamma_{\xi} I & 0 & S\\
\nabla_x f & 0 & 0 & 0\\
\nabla_x g_{\mathcal A} & S & 0 & 0
},
\end{equation*}
where $H
:=
\nabla^2_{xx}
\left(
\ell(x)+y_f^\top f(x)+y_{\mathcal A}^\top g_{\mathcal A}(x)
\right).$

To solve this linear system, we distinguish two cases.

\paragraph{No slack variables $(\delta_\xi,\xi)=(0,0)$}
When slack variables are disabled, the active inequality constraints are
\[
g_{\mathcal A}(x)=0.
\]
The KKT conditions reduce to
\begin{subequations}
\label{eq:bwd:noslack_kkt}
\begin{align}
\nabla_x \ell(x)
+\nabla_x f(x)^\top y_f
+\nabla_x g_{\mathcal A}(x)^\top y_g
&=0,\\
f(x)&=0,\\
g_{\mathcal A}(x)&=0,
\end{align}
\end{subequations}
with $F(w,\theta)=0$,  $
w=(x,y)$, and 
\begin{equation}
\label{eq:bwd:noslack_kkt_matrix}
\frac{\partial F}{\partial w}
=
\AverageSmallMatrix{
H & \nabla_x f^\top & \nabla_x g_{\mathcal A}^\top\\
\nabla_x f & 0 & 0\\
\nabla_x g_{\mathcal A} & 0 & 0
}.
\end{equation}
The gradient is then
\begin{align}\label{eq:loss_gradient}
\hspace{-2mm}
\nabla_\theta \mathcal L^\top 
&= 
-\frac{\partial F}{\partial \theta}^\top
\left(
\left[\frac{\partial F}{\partial w} \right]^{-1}
\begin{bmatrix}
\nabla_x\mathcal L\\
0
\end{bmatrix}
\right)
= 
-\frac{\partial F}{\partial \theta}^\top
\begin{bmatrix}
\eta_x\\
\eta_y
\end{bmatrix}
\end{align}
where $
\AverageSmallMatrix{
\eta_x\\
\eta_y
}$ solves the linear system $
\frac{\partial F}{\partial w}
\AverageSmallMatrix{
\eta_x\\
\eta_y
}
=
\AverageSmallMatrix{
\nabla_x\mathcal L\\
0
}
$. \\

\paragraph{Slack variables ($\delta_\xi=1$)}
We first eliminate the slack adjoints $\eta_\xi$.  From the last and second block rows of
\eqref{eq:bwd:slack_adjoint},
\begin{equation}
\label{eq:bwd:slack:eta_xi_yg}
\eta_\xi=-S\nabla_x g_{\mathcal A}\,\eta_x, 
\quad 
\eta_{y_g}
=
S\nabla_{\xi_{\mathcal A}}\mathcal L
+
\gamma_{\xi} \nabla_x g_{\mathcal A}\,\eta_x. 
\end{equation}
By substituting~\eqref{eq:bwd:slack:eta_xi_yg} into the first block row  of
\eqref{eq:bwd:slack_adjoint}, we obtain the reduced linear system
\begin{equation}
\label{eq:bwd:slack_reduced_system}
\scalemath{0.9}{\begin{bmatrix}
H{+}\gamma_{\xi} \nabla_x g_{\mathcal A}^\top\nabla_x g_{\mathcal A}
&
\nabla_x f^\top\\
\nabla_x f
&
0
\end{bmatrix}
\begin{bmatrix}
\eta_x\\
\eta_{y_f}
\end{bmatrix}
=
\begin{bmatrix}
\nabla_x\mathcal L
{-}
\nabla_x g_{\mathcal A}^\top S\nabla_{\xi_{\mathcal A}}\mathcal L\\
0
\end{bmatrix}}.
\end{equation}
After solving \eqref{eq:bwd:slack_reduced_system},  $(\eta_\xi,\eta_{y_g})$ are
recovered via \eqref{eq:bwd:slack:eta_xi_yg}.

If the slack penalty $\gamma_{\xi}$ is also a parameter to differentiate, since the only explicit dependence of
$F$ on $\gamma_{\xi}$ is through the stationarity condition
$\gamma_{\xi}\xi_{\mathcal A}+Sy_g=0$,
\begin{equation}
    \frac{\partial \mathcal L}{\partial \gamma_{\xi}}
=
-\xi_{\mathcal A}^\top \eta_\xi
=
\xi_{\mathcal A}^\top S\nabla_x g_{\mathcal A}\,\eta_x.
\end{equation}

\paragraph{Solving the linear systems in the backward pass}
The backward pass requires solving a linear system that shares the time-induced sparsity of the forward primal updates, including: a block-tridiagonal Hessian, a block-bidiagonal equality Jacobian $\nabla_x f$, and a block-diagonal active inequality Jacobian $\nabla_x g_{\mathcal A}$. However, the forward and backward linear-system solves diverge in the structure of the assembled KKT matrix. \vspace{-0.25cm}
\begin{multicols}{2}
\begin{center}
\textit{Forward Pass}
\end{center}
\[
\begin{bmatrix}
\star & \star\\
\star & -\rho^{-1}I
\end{bmatrix}
\]

\columnbreak

\begin{center}
\textit{Backward Pass}
\end{center}
\[
\begin{bmatrix}
\star & ~~\star~~\\
\star & ~~0~~
\end{bmatrix}
\]
\end{multicols}

In the forward pass, the $-\rho^{-1}I$ block is derived from the ADMM augmented Lagrangian, while in the backward-pass matrix $\partial F/\partial w$ \eqref{eq:bwd:LS}, the bottom right block is  zero.
The backward pass differentiates the KKT stationarity conditions at the forward-pass converged solution, at which the penalty over the inequality constraints via $\rho$ has no role. 

Rather than attempting a Schur reduction on the backward-pass linear system, we assemble the full sparse linear system in CSR (Compressed Sparse Row) format and solve it directly using \texttt{cuDSS} in general-symmetric mode, which handles indefiniteness natively via $LDL^\top$ decomposition. This implementation entails our default and fastest GPU backward-solver backend, but we also provide a pure-JAX reference, (i.e., \texttt{direct\_jax\_dense}) useful for debugging and CPU-only testing.    Table~\ref{tab:backends} identifies the backends that we make available through our implementation.

We also explored solving the backward pass iteratively by formulating a reduced adjoint QP whose optimality conditions approximate the sensitivity system. This QP takes the form:
\begin{subequations}
\begin{align}
\min_{x}\quad&
\Big(
\tfrac12 \tilde x^\top \left[\nabla_{xx}L 
{+}
\gamma_{\xi} \nabla_x \Delta g_{\mathcal A}^\top
\nabla_x \Delta g_{\mathcal A}\right] \tilde x
\\ \nonumber
&\qquad -
(\tfrac{\partial \mathcal{L}}{\partial x}
-
\nabla_x \Delta g_{\mathcal A}^\top
\tfrac{\partial \mathcal{L}}{\partial \xi_{\mathcal A}})^\top \tilde x
\big)
\\
\text{s.t.}\ \ 
&\nabla_x f\, \tilde x=0, % 
\end{align}
\end{subequations}
where the active inequality constraints are absorbed into the Hessian via the penalty term $\gamma_{\xi}$, producing a convex equality-only QP that reuses the forward ADMM infrastructure directly.
This approach recovers accurate gradients when the ADMM iterates are converged tightly, but introduces a small bias proportional to the convergence tolerance and the active-set approximation error. In contrast, the direct-backward solve via \texttt{cuDSS} yields the exact sensitivity in a single factorization. Our open-sourced code enables testing these different backward-solve strategies, by choosing between direct- and ADMM-based solver backends (Table~\ref{tab:backends}).

\begin{table}[!t]
\centering
\caption{QP solver backends.
All backends are available for both passes
unless marked \textsuperscript{$\dagger$} (backward only).}
\label{tab:backends}
\begin{tabular}{ll}
\toprule
Backend & Schur solver \\
\midrule
\texttt{admm\_jax\_loop\_pcg}       & PCG (JAX) \\
\texttt{admm\_jax\_loop\_pcg\_ffi}  & PCG (FFI) \\
\texttt{admm\_jax\_loop\_cudss\_ffi}& cuDSS (FFI) \\
\texttt{admm\_jax\_loop\_jax\_dense}& Dense (JAX) \\
\texttt{admm\_fused\_pcg}           & PCG (fused) \\
\texttt{admm\_fused\_cudss}          & cuDSS (fused) \\
\midrule
\texttt{direct\_jax\_dense}\textsuperscript{$\dagger$}  & Direct KKT (JAX) \\
\texttt{direct\_cudss\_ffi}\textsuperscript{$\dagger$}  & Direct KKT (cuDSS) \\
\bottomrule
\end{tabular}
\end{table}

\begin{figure*}[t]
\centerline{\includegraphics[width=0.95\linewidth]{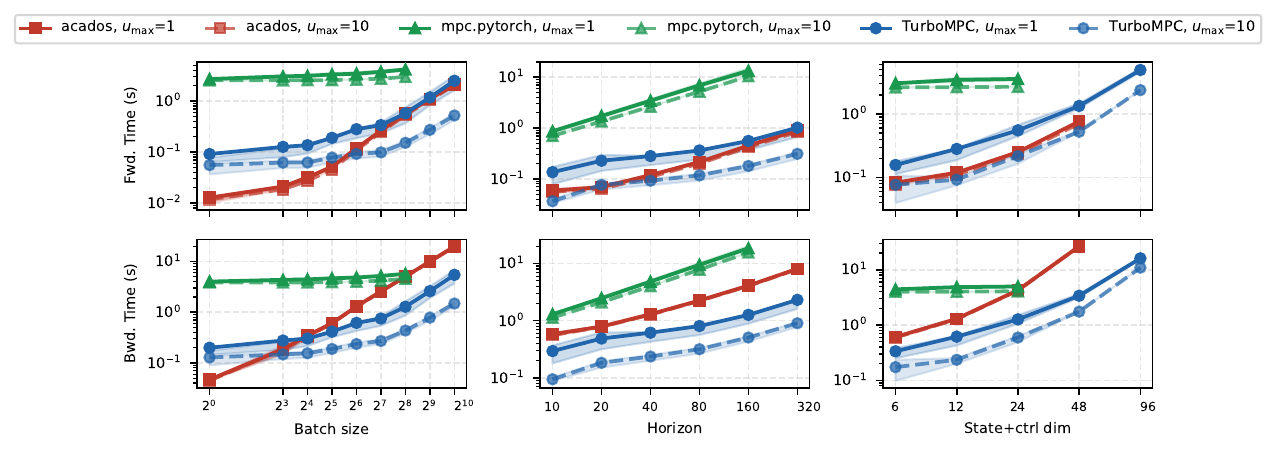}}
\vspace{-10pt}
\caption{Linear-system RL ($\texttt{tol}{=}10^{-3}$): mean solve time vs.\ batch size (left), planning horizon (center), state $+$ control dimension (right) for \solverName~(blue), \texttt{acados} (red), and \texttt{mpc.pytorch} (green) at $u_{\max}\in\{1,10\}$. \solverName's GPU advantage grows with batch size, problem size, and horizon.}
\label{fig:bench:ls-rl-1e-3}
\vspace{-10pt}
\end{figure*}

\begin{figure}[!t]
\centering
\includegraphics[width=0.95\linewidth]{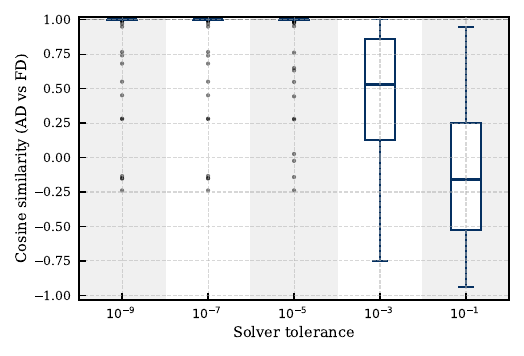}
\vspace{-10pt}
\caption{Linear-system RL: Cosine similarity between computed gradients and finite differencing across solver tolerances.}
\label{fig:gradient_accuracy}
\vspace{-10pt}
\end{figure}

\subsection{Linear-System RL Scaling}\label{appendix:linear}
This section provides additional details regarding scalability and gradient accuracy. 

First, concerning scalability, Fig.~\ref{fig:bench:ls-rl-1e-3} shows scalability results with tolerance $\texttt{tol}=10^{-3}$. 
Compared to \texttt{mpc.pytorch} in this case, the gain reaches $24.0\times$ and $14.7\times$ speedups for the forward and backward passes with the nominal configuration and $u_{max}=1.0$. The highest gain achieved corresponds to a $58.1\times$ speedup with $u_{max}=10.0$ for the same nominal configuration. On the other hand, compared to \texttt{acados}, \solverName~starts to consistently outperform \texttt{acados} at batch sizes $8$--$32$ on the backward pass, achieving $3.6\times$ and $13.3\times$ gains for the backward pass at batch size $1024$, with $u_{max}=1.0$ and $u_{max}=10.0$, respectively. The highest gain compared to \texttt{acados} is a $15.1\times$ speedup for the backward pass for $(B,N,n_x,n_u)=(64,160,32,16)$ with $u_{max}=10.0$. 
These results demonstrate the superior speedup performance that \solverName~can achieve at deployment, when more computational effort lies on the forward than the backward pass. 

Concerning the backward pass, we evaluate the accuracy of the gradients computed by \solverName~against finite differences approximations depending on the solver tolerance. Fig.~\ref{fig:gradient_accuracy} reports the results for the nominal configuration with $u_{max}=1.0$ and statistics computed over 100 problem instances. The gradient accuracy increases with tighter tolerances. The outlier values can show low-primal feasibility solution due to the random seeds, or gradient discontinuities due to the solution landing on a constraint boundary that turns the active set ambiguous. Nonetheless, as noticed before, a tolerance below $10^{-5}$ produces accurate gradients. This analysis motivates future work to improve gradients computation at coarser tolerances.

\subsection{Humanoid Balancing Imitation Learning} \label{appendix:humanoid}
\subsubsection{State and control}
The state $x_t = (p_t, \Theta_t, v_t, \omega_t) \in \R^{12}$ comprises the CoM position $p_t \in \R^3$, the $\rm{ZYX}$ Euler angles (roll, pitch, yaw) $\Theta_t = (\phi_t, \theta_t, \psi_t) \in \mathbb{R}^3$, and the linear and angular velocities, $v_t \in \R^3$ and $\omega_t \in \R^3$, respectively.

The control $u_t=(f_{L,t}, f_{R,t}) \in \R^6$ includes the 
left, $f_{L,t}$, and right, $f_{R,t}$, foot contact forces in the world frame.

\subsubsection{Single Rigid Body (SRB) dynamics}
The SRB dynamics are governed by:
\begin{subequations} \label{eq:SRB_dyns}
\begin{align}
\dot{p}_t &= v_t, \label{eq:pos}\\
\dot{\Theta}_t &= T(\Theta_t) \omega_t, \label{eq:euler}\\
\dot{v}_t &= \frac{1}{m}(f_{L,t} + f_{R,t}) + g, \label{eq:vel}\\
\dot{\omega}_t &= I^{-1}\left(r_L \times f_{L,t} + r_R \times f_{R,t} - \omega_t \times (I\omega_t)\right), \label{eq:ang}
\end{align}
\end{subequations}
where $m = 33.34$ kg is the total robot mass, $g = (0, 0, -9.81)$ m/s$^2$ is the gravity acceleration vector, $I \in \R^{3 \times 3}$ is the body-frame inertia tensor, and $r_L, r_R \in \R^3$ are contact position vectors from CoM to the corresponding feet.

The ZYX convention relates body-frame angular velocity $\omega_t$ to Euler angle rates via:
\begin{equation}
T(\Theta_t) = \begin{bmatrix}
  1 & \sin\phi_t \tan\theta_t & \cos\phi_t \tan\theta_t \\
  0 & \cos\phi_t & -\sin\phi_t \\
  0 & \sin\phi_t / \cos\theta_t & \cos\phi_t / \cos\theta_t
\end{bmatrix}.
\end{equation}

This parameterization avoids 3D rotation matrix Lie group complexity, and has a singularity at $\theta_t = \pm\pi/2$ (gimbal lock) that does not occur during balance recovery. The parameterization is a deliberate simplification, acknowledging that unit quaternions (or another (SO(3))-consistent representation) would be the representation of choice for robust full locomotion with large rotations. 

We use a 4th-order Runge-Kutta (RK4) scheme with $\Delta t=0.025~\rm{secs.}$:
$x_{t+1} = \Phi(x_t, u_t),$
where $\Phi(\cdot)$ is a function of the continuous-time SRB dynamics in~\eqref{eq:SRB_dyns} via the RK4 coefficients.

\subsubsection{Optimal Control Problem} 
We solve the   OCP
\begin{subequations}
    \begin{align}
    \min_{x,u}\ & 
    \sum_{t=0}^{N} 
      \delta x_t^\top Q(\theta)~\delta x_t + \delta u_t^\top R~\delta u_t
 \\ 
    \text{s.t.} \quad &x_0 =x_{\rm init},\\  
    &x_{t+1} = \Phi(x_t, u_t) ~~\quad \forall t = 0, \ldots, N-1,\\  
    &u_{\min} \le u_t \le u_{\max} \quad \forall t = 0, \ldots, N.
    \end{align}
\end{subequations}
where $\delta x_t=x_t - x_{\text{ref}}$, $\delta u_t=u_t - u_{\text{ref}}$, and $\theta\in\R^{12}$ represents the learnable cost weights used to parameterize the state cost matrix $Q(\theta)$, $R=\mathrm{diag}(10^{-3},\ldots,10^{-3})\in\R^{6\times6}$ is a control cost matrix, and $N = 20$ is the horizon length.

\subsubsection{Learnable cost weights}
To ensure positive semi-definiteness, we parameterize the weights in log-space:
\begin{align}
Q(\theta) &= \text{diag}(\exp(\theta_1), \ldots, \exp(\theta_{12})) \in \mathbb{R}^{12 \times 12}.
\end{align}
The log-space parameterization ensures positivity by construction and allows the optimizer to work in an unconstrained search space.

\subsubsection{Friction cone as box bounds}
Friction cone constraints are approximated as independent box bounds:
\begin{subequations}
    \begin{align}
\label{eq:box}
u_{\min} &= (-240, -240, 0, -240, -240, 0) \text{ (N)},\\
u_{\max} &= (240, 240, 400, 240, 240, 400) \text{ (N)},
\end{align}
\end{subequations}
which enforces:
\begin{itemize}
  \item Horizontal forces: $|f_x|, |f_y| \le 240$ N
  \item Vertical forces: $0 \le f_z \le 400$ N
\end{itemize}
The box structure avoids second-order cone constraints and is directly compatible with \solverName.

\subsubsection{WBC approximation}
We implement a SRB MPC that outputs desired ground reaction forces $f_{L,t}, f_{R,t}$ mapped to joint torques using a static whole-body inverse-dynamics approximation on the full MuJoCo model:
\begin{equation}
\tau_t = \left(h(q_t, 0) - J_{\text{feet}}(q_t)^\top f_{\text{des},t}\right)_{\text{ID}}
\end{equation}
where $f_{\text{des},t} = (f_{L,t}, f_{R,t})$, $h$ is the MuJoCo bias-force vector (gravity + Coriolis + centrifugal), and $J_{\text{feet}}$ is the stacked translational foot Jacobian. The operator $(\cdot)_{\text{ID}}$ selects actuated joints (29 DoF) from generalized coordinates (35 DoF). This setting corresponds to a quasi-static WBC assumption with $\dot{q}=0$ and $\ddot{q}=0$, and it is intentionally used as a bridge from centroidal MPC to full-body torque commands during balance-recovery experiments. A dynamic acceleration-level WBC is deferred to a walking-phase implementation as future work.

\subsubsection{MuJoCo validation protocol}
We validate the SRB-to-WBC pipeline in MuJoCo with three modes:
\begin{itemize}
  \item Kinematic replay: set floating-base pose from SRB states, no physics stepping.
  \item Open-loop physics: apply precomputed WBC torques and run forward dynamics.
  \item Closed-loop physics: apply
  \begin{equation}
  \tau_{\text{applied},t} = \tau_{\text{WBC},t} + K_p(q_{\text{des}}-q_t) + K_d(\dot{q}_{\text{des}}-\dot{q}_t),
  \end{equation}
  with $q_{\text{des}}=0$, $\dot{q}_{\text{des}}=0$ for nominal standing. $K_p$ and $K_d$ are diagonal, joint-group-dependent PD feedback gains:
\begin{subequations}
    \begin{align}
K_p &= \mathrm{diag}(\underbrace{200,\ldots,200}_{12\ \text{leg joints}},\underbrace{100,100,100}_{3\ \text{waist joints}},\underbrace{50,\ldots,50}_{14\ \text{arm joints}}),\\
K_d &= \mathrm{diag}(\underbrace{10,\ldots,10}_{12\ \text{leg joints}},\underbrace{5,5,5}_{3\ \text{waist joints}},\underbrace{2,\ldots,2}_{14\ \text{arm joints}}),
\end{align}
\end{subequations}

with units N$\cdot$m/rad and N$\cdot$m$\cdot$s/rad, respectively.
\end{itemize}

The open-loop mode validates short-horizon feasibility but drifts over longer horizons due to model mismatch and lack of feedback. The closed-loop mode reduces  drift and is used as a practical validation mode. Future work includes  using time-varying contact schedules and dynamic WBC for walking.

\subsection{Neural Network Training}
\label{app:pointmass-rl}
We use \solverName \,as a differentiable MPC policy with a neural-network cost function for a constrained
3D point-mass tracking task. The state is
$x=(p,v)\in\mathbb{R}^{6}$, where
$p\in\mathbb{R}^{3}$ is the position and $v\in\mathbb{R}^{3}$ is the velocity.
The control is the thrust vector $u\in\mathbb{R}^{3}$. The dynamics are
a unit-mass double integrator discretized with forward Euler at
$\Delta t=0.1$, with thrust box constraints $\|u_t\|_\infty\le 5$ N. The MPC planning
horizon is $N=8$, and the closed-loop RL training rollout horizon is $H=8$.

The reference is a periodic zig-zag trajectory defined by six waypoints.
Each waypoint is held for four control steps. After reaching the last
waypoint, the reference reverses through the waypoint sequence. This gives the
$40$-step cycle in Table~\ref{tab:pointmass-reference}. 
At each MPC call, the reference window contains the next $N+1$ waypoint
targets from the current phase.

\begin{table}[t]
    \centering
    \caption{Periodic point-mass reference trajectory over one cycle.}
    \label{tab:pointmass-reference}
    \begin{tabular}{c c c c}
        \hline
        Cycle segment & Waypoint & Steps $i$ & $p_i^{\mathrm{ref}}=(x,y,z)$ m \\
        \hline
        0 & 0 & $0$--$3$   & $(\phantom{-}0.10,\ 0.00,\ 0.00)$ \\
        1 & 1 & $4$--$7$   & $(-0.10,\ 0.05,\ 0.05)$ \\
        2 & 2 & $8$--$11$  & $(\phantom{-}0.10,\ 0.10,\ 0.10)$ \\
        3 & 3 & $12$--$15$ & $(-0.10,\ 0.15,\ 0.15)$ \\
        4 & 4 & $16$--$19$ & $(\phantom{-}0.10,\ 0.10,\ 0.10)$ \\
        5 & 5 & $20$--$23$ & $(-0.10,\ 0.05,\ 0.05)$ \\
        6 & 4 & $24$--$27$ & $(\phantom{-}0.10,\ 0.10,\ 0.10)$ \\
        7 & 3 & $28$--$31$ & $(-0.10,\ 0.15,\ 0.15)$ \\
        8 & 2 & $32$--$35$ & $(\phantom{-}0.10,\ 0.10,\ 0.10)$ \\
        9 & 1 & $36$--$39$ & $(-0.10,\ 0.05,\ 0.05)$ \\
        \hline
    \end{tabular}
\end{table}

The learned parameters $\theta$ are the weights of a two-layer neural network with
one hidden layer,
\(
    6 \rightarrow 8 \rightarrow 81.
\)
The hidden activation is $\tanh$.  The first-layer weights are initialized i.i.d.\ from a zero-mean Gaussian with standard deviation $0.5$, all biases and the output layer weights are initialized to zero.   Thus, at initialization the network outputs zero log-multipliers and the controller recovers the default MPC cost prior $(Q_t^\theta,R_t^\theta ) = (I_6,10^{-3} I_3)$ for all stages $t$. The network output is reshaped as a $(N+1)\times 9$ matrix, providing time-varying cost weights.

At closed-loop phase $i$, the network evaluates the current state $x_i$
and produces one row
$\eta_t=(\eta^p_t,\eta^v_t,\eta^u_t)\in\mathbb{R}^{9}$ for
each MPC stage $t=0,\ldots,N$, with
$\eta^p_t,\eta^v_t,\eta^u_t\in\mathbb{R}^3$. These outputs
assemble the MPC cost matrices
\begin{align}
    Q_t^\theta
    &=
    \mathrm{diag}\!\left(
        \exp(\eta^p_t),
        \exp(\eta^v_t)
    \right)\in\mathbb{R}^{6\times 6},\\
    R_t^\theta
    &=
    10^{-3}\,\mathrm{diag}\!\left(\exp(\eta^u_t)\right)
    \in\mathbb{R}^{3\times 3}.
\end{align}
The MPC problem solved at that closed-loop step is
\begin{subequations}
\begin{align}
\mathop{\min}_{(x,u)}\ \ \,
&
\mathrlap{
\sum_{t=0}^{N}
(x_t\,{-}\,x^{\textrm{ref}}_{i+t})^\top Q_t^\theta
(x_t\,{-}\,x^{\textrm{ref}}_{i+t})
\,{+}\,u_t^\top R_t^\theta u_t,
}
\\
\text{s.t.}\ \ \ 
&x_0
=
x_i,
\\
&p_{t+1}
=
p_t+\Delta t\,v_t,
&&t=0,\ldots,N-1,
\\
&v_{t+1}
=
v_t+\Delta t\,u_t,
&&t=0,\ldots,N-1,
\\
&\|x_t\|_\infty
\le 5,
\ 
\|u_t\|_\infty\le 5,
&&t=0,\ldots,N,
\end{align}
\end{subequations}
where $x^{\textrm{ref}}_{i+t}\,{=}\,(p_{i+t}^{\mathrm{ref}},0,0,0)$
is the phase-aligned position reference with zero velocity reference.
The policy applies only the first control
$\pi_\theta(x_i)=u_0^\star$ before resolving the MPC problem at the
next state. The exponential parameterization keeps all MPC cost weights
positive while allowing unconstrained optimization over $\theta$.

Training minimizes the expected closed-loop tracking loss
\begin{equation}
\label{eq:pointmass-apg-loss}
\resizebox{0.88\columnwidth}{!}{$
\mathcal{L}(\theta)
=
\mathbb{E}_{x_0,i_0}
\left[
\frac{1}{H}\sum_{t=0}^{H-1}
\|p_{t+1}-p^{\mathrm{ref}}_{i_0+t+1}\|_2^2
+
10^{-3}\|u_t\|_2^2
\right]
$}.
\end{equation}
The phase $i_0$ is sampled uniformly over one reference cycle. Initial
states $x_0$ are sampled near the reference: the initial position is
the waypoint at phase $i_0$ plus uniform perturbation in
$[-0.03,0.03]^3$, and the initial velocity is sampled uniformly in
$[-0.05,0.05]^3$. We optimize \eqref{eq:pointmass-apg-loss} with Adam
using batch size $B=4$, learning rate $3\times10^{-3}$, and $500$
gradient steps.

We use analytic policy gradients to train the neural network, which flow through the closed-loop
rollout, with derivatives propagated through each MPC solution used by
the policy.

For evaluation, the system starts on
the first reference waypoint with zero velocity and is simulated for
$80$ closed-loop steps. The reported metric is the position RMSE against
the phase-aligned reference. Training takes $30.9$ minutes on an RTX 5090,
with a median training step time of $3.2$ seconds. The tracking RMSE is reduced from $1.20\times10^{-1}$ m at initialization to
$8.33\times10^{-2}$ m after $500$ gradient steps. % 

\bibliographystyle{inc/IEEETran}
\bibliography{inc/main}

\end{document}